\documentclass{article}

\usepackage[preprint]{neurips_2026}
\usepackage{subcaption}
\usepackage{pifont}
\usepackage[most]{tcolorbox}
\usepackage{xcolor}

\definecolor{mygray}{HTML}{f6f6f7}
\definecolor{myp}{HTML}{dddfeb}

\tcbset{
  promptbox/.style={
    enhanced,
    breakable,
    colback=white,
    colframe=myp,
    boxrule=0.8pt,
    arc=2mm,
    left=1.2mm,
    right=1.2mm,
    top=1.2mm,
    bottom=1.2mm,
    fonttitle=\bfseries,
    coltitle=black,
    colbacktitle=mygray,
  }
}
\usepackage{amsmath}
\usepackage[utf8]{inputenc} % allow utf-8 input
\usepackage[T1]{fontenc}    % use 8-bit T1 fonts
\usepackage{hyperref}       % hyperlinks
\usepackage{url}            % simple URL typesetting
\usepackage{booktabs}       % professional-quality tables
\usepackage{amsfonts}       % blackboard math symbols
\usepackage{nicefrac}       % compact symbols for 1/2, etc.
\usepackage{microtype}      % microtypography
\usepackage{xcolor}         % colors
\usepackage{graphicx} 
\usepackage{subcaption}
\usepackage{wrapfig}

\usepackage[table]{xcolor}
\usepackage{colortbl}
\usepackage[most]{tcolorbox}
\usepackage{fontawesome5}
\usepackage{enumitem}
\definecolor{topone}{HTML}{89E0CD}
\definecolor{toptwo}{HTML}{A9EADF}
\definecolor{topthree}{HTML}{C8F3EA}

\definecolor{bottomone}{HTML}{C7AEBB}
\definecolor{bottomtwo}{HTML}{D8C4CE}
\definecolor{bottomthree}{HTML}{E8DDE3}

\definecolor{mygreen}{HTML}{89E0CD}
\definecolor{myred}{HTML}{EFC4CD}

\newcommand{\bestA}[1]{\cellcolor{topone}\textbf{#1}}
\newcommand{\bestB}[1]{\cellcolor{toptwo}#1}
\newcommand{\bestC}[1]{\cellcolor{topthree}#1}

\newcommand{\worstA}[1]{\cellcolor{bottomone}#1}
\newcommand{\worstB}[1]{\cellcolor{bottomtwo}#1}
\newcommand{\worstC}[1]{\cellcolor{bottomthree}#1}

\definecolor{mygreenn}{HTML}{ae5a41}
\definecolor{myredd}{HTML}{559e83}
\definecolor{mygrayy}{HTML}{D0D5D8}
\definecolor{myorangee}{HTML}{9f709e}
\definecolor{mypurplee}{HTML}{00c2c7}
\definecolor{myneww}{HTML}{3bb99d}
\definecolor{mydis}{HTML}{a2798f}

\hypersetup{
    colorlinks=true,
    linkcolor=mypurplee,  
    citecolor=mygreenn,   
    urlcolor=myorangee     
}
\title{ViMU: \\Benchmarking Video Metaphorical Understanding}

\author{
   Qi Li  \quad Xinchao Wang\thanks{Corresponding Author} \\
  National University of Singapore \\
  \texttt{liqi@u.nus.edu} \quad \texttt{xinchao@nus.edu.sg}
}

\begin{document}

\maketitle

\vspace{-2.5em}
\begin{center}
\small
\href{https://liqiiiii.github.io/Video-Metaphorical-Understanding/}{\faGlobe\ Project Page}
\quad\quad
\href{https://github.com/LiQiiiii/Video-Metaphorical-Understanding}{\faGithub\ GitHub}
\quad\quad
\href{https://huggingface.co/datasets/LIQIIIII/ViMU}{\faDatabase\ Dataset}
\end{center}

\begin{abstract}
Any new medium, once it emerges, is used for more than the transmission of overt content alone. The information it carries typically operates on two levels: one is the content directly presented, while the other is the subtext beneath it—the implicit ideas and intentions the creator seeks to convey through the medium. Likewise, since video technologies became widely adopted, video has served not only as a powerful tool for recording and communicating visual information, but also as a vehicle for emotions, attitudes, and social meanings that are often difficult to articulate explicitly. Thus, the true meaning of many videos does not reside solely in what is shown on screen; it is often embedded in context, style of expression, and the viewer’s social experience. Some forms of such video subtext are humorous, while others carry irony, mockery, or criticism. These implicit meanings can also be interpreted very differently across cultural backgrounds and social groups. However, most existing video understanding models still focus primarily on literal visual comprehension, such as recognizing objects, actions, or temporal relations, and lack a systematic ability to understand the metaphorical, ironic, and social meanings embedded in videos. To bridge this gap, we introduce \texttt{ViMU} (\underline{Vi}deo \underline{M}etaphorical \underline{U}nderstanding), the first benchmark designed to systematically evaluate the subtext understanding capabilities of frontier models in videos. ViMU assesses whether video understanding models can go beyond literal perception to infer implicit meaning, rhetorical devices, social signals, target subjects, and culturally grounded subtext, while grounding their interpretations in multimodal evidence and answering both open-ended and multiple-choice questions. Importantly, all questions are designed to be hint-free, ensuring that no key evidence is disclosed to models before answering. Extensive experiments show that most frontier models, including closed-source ones, achieve below 50\% overall performance. We further conduct fine-grained analyses to uncover distinctive model behaviors. \textcolor{mydis}{\textbf{Disclaimer:} This paper contains potentially offensive and harmful content.}
\end{abstract}

\begin{flushright}
    \begin{minipage}{0.45\textwidth} %
        \flushright
        ``The most important thing in communication is hearing what isn't said.'' \\
        \vspace{0.5em}
        \hrule 
        \vspace{0.5em}
        \textit{- Peter Drucker}
    \end{minipage}
\end{flushright}

\section{Introduction}

Recent advances in large language models have enabled the integration of rich real-world information, including videos, into model representations~\cite{achiam2023gpt,guo2025seed1,bai2025qwen3,yang2025qwen3,team2024gemma,li2026vid,li2026cola,wang2025towards,yu2025discrete}. Consequently, video understanding models have become effective for tasks such as visual grounding and causal reasoning~\cite{fu2025video,zhou2025mlvu,wang2025lvbench}.
Yet these forms of understanding remain largely confined to the surface-visible content. Put simply, directly observable content explains how an event unfolds, but not what it ultimately means, as such meaning often lies in the underlying social subtext\footnote{As Roland Barthes notes in his book \textit{Mythologies}, "myth is a second-order semiological system"~\cite{leak1994barthes,wiki_mythologies}, in which literal content serves as the basis for a secondary layer of cultural or ideological meaning.}: the deeper layer that maps an event onto broader social meanings, values, and collective attitudes. Together, the visible content and its subtext constitute the full depth of video understanding~\cite{leak1994barthes,hall2019encoding,kress2020reading}.

\begin{figure}[t]
    \centering
    \includegraphics[width=\textwidth]{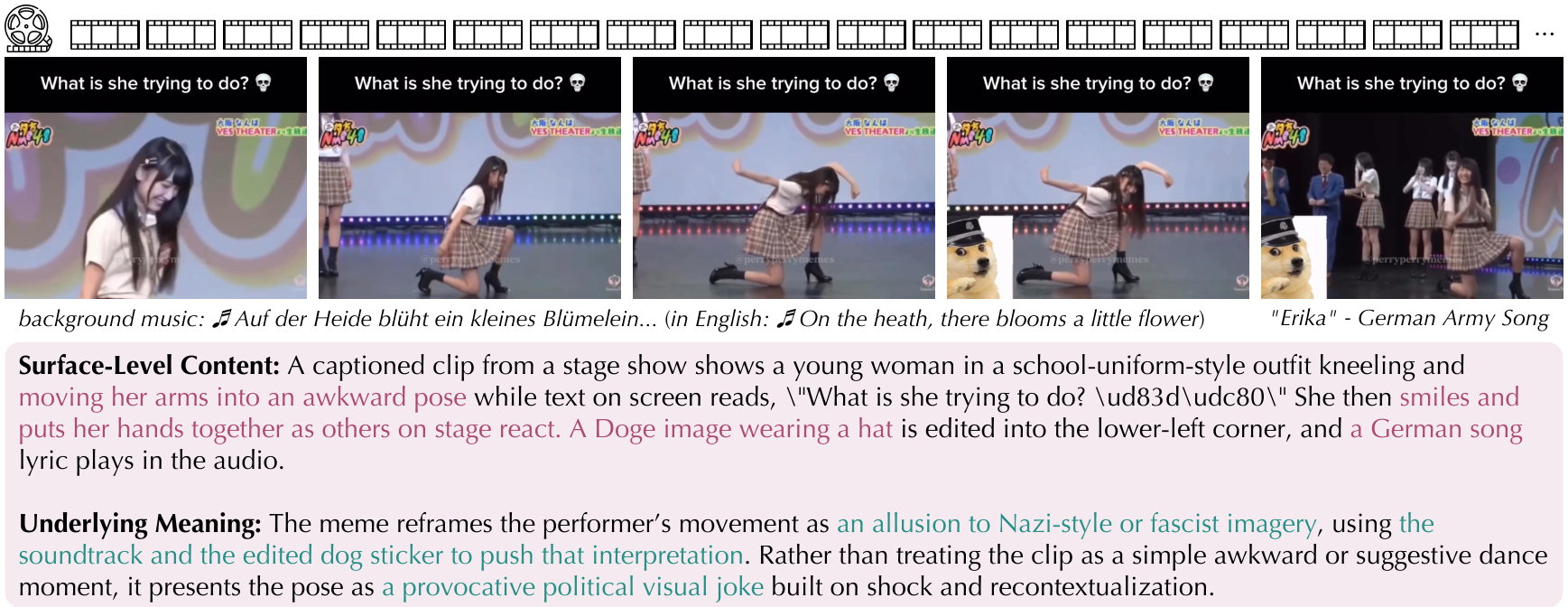}
    % \vspace{-5mm}
    \includegraphics[width=\textwidth]{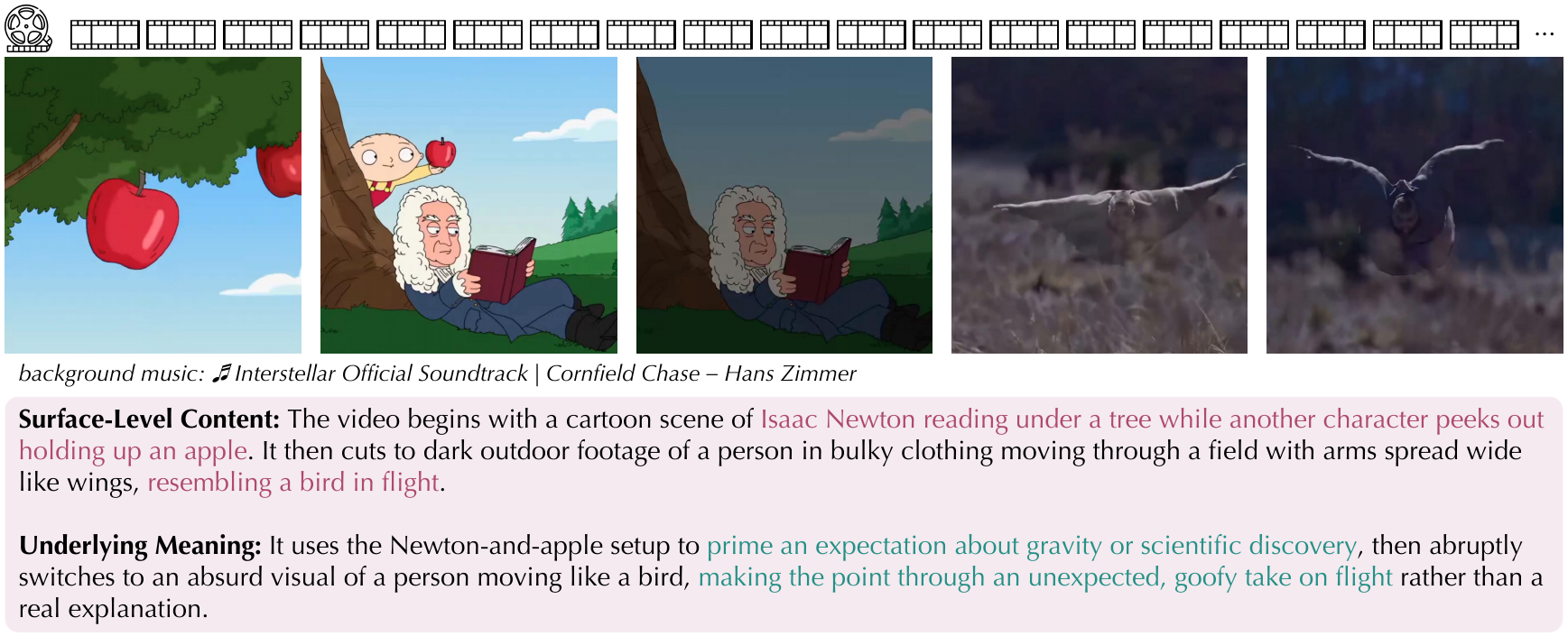}
    
    \caption{Examples illustrating the large gap between observable content and underlying subtext in videos. In the top example, the video appears to show a girl dancing on a reality show, while its implied meaning alludes to Nazi symbolism. In the bottom example, the video appears to show a child catching an apple above Newton and a strange flying scene, while its underlying joke is that the apple missing Newton led to a setback in the development of physics.}
    \vspace{-5mm}
    \label{fig:pipeline}
\end{figure}

As illustrated in Figure~\ref{fig:pipeline}, the gap between observable content and underlying subtext can be substantial.
In such a case, understanding the video requires more than recognizing objects, actions, or temporal structure, which are typically emphasized in prior works~\cite{fu2025video,zhou2025mlvu,wang2025lvbench,xiao2024can,chen2024mecd,li2024mvbench}. It demands integrating multimodal evidence, recovering culturally situated references, and inferring the creator’s communicative intent beyond what is explicitly shown. Existing evaluations left far behind for such subtext interpretation in videos.
Most existing benchmarks fall short in three ways: (i) targeting implicit reasoning over hidden spatial, physical, or interactional relations rather than socially grounded meanings~\cite{swetha2026vrrqavisualrelationalreasoning,chen2025looking}; (ii) focusing only on narrower phenomena such as non-verbal humor~\cite{shi2025v}; or (iii) relying on multiple-choice formats whose options may expose plausible subtext hypotheses~\cite{jiang2026avmeme}. These settings do not fully capture genuine hint-free inference over socially grounded video meaning.

To fill this gap, we introduce \texttt{ViMU}, a benchmark specifically designed to evaluate whether models can move beyond observable content to recover the underlying subtext of videos. In particular, ViMU requires models to infer implicit meaning in a hint-free manner, without being told in advance which socio-cultural cues are relevant.
To achieve this, we build ViMU through a meticulous curation process involving multiple rounds of annotation and filtering by advanced closed-source models and human experts. This procedure is designed not only to ensure task difficulty and a genuinely hint-free evaluation setting, but also to maintain broad coverage of diverse rhetorical mechanisms and social value signals. Finally, we obtain a high-quality dataset of 588 videos with 2,352 questions across four tasks, covering both open-ended and multiple-choice questions.

We extensively investigate 16 popular MLLMs with ViMU, which brings in several critical insights. Firstly, \textit{video metaphorical understanding remains a technically challenging problem for the existing MLLMs}. Even the most advanced closed-source models achieve below 50\% average performance across the four tasks. Secondly, \textit{many models systematically over-predict generic or safer categories while under-predicting more implicit or socially coded ones}, suggesting a shared tendency to favor more accessible interpretations over deeper subtextual inference.
Thirdly, we observe \textit{a clear mismatch between general video understanding and metaphorical video understanding}: models that excel on conventional video understanding task do not necessarily perform best on our tasks.
In addition to the overall conclusion, individual tasks enable fine-grained analysis in each specialized aspects. Therefore, we anticipate the benchmark to assist in improving MLLMs’ video metaphorical understanding capabilities by providing insights into their current strengths and weaknesses.

\section{Related Work}

\paragraph{Reasoning beyond explicit visual evidence.}
Some recent work has moved beyond explicit-evidence-centric VideoQA by requiring models to infer answers from indirect or partially unavailable cues. I-VQA~\cite{chen2025looking} studies settings where explicit visual evidence is missing and answers must be inferred from context, building on related work in visual commonsense and context-based reasoning such as VisualCOMET~\cite{park2020visualcomet}, Video2Commonsense~\cite{fang2020video2commonsense}, and causal video reasoning methods like MECD~\cite{chen2024mecd} and MECD+~\cite{chen2025mecdplus}. VRR-QA~\cite{swetha2026vrrqavisualrelationalreasoning} further focuses on implicit relational reasoning across frames when key relations are not directly co-visible. While these benchmarks go beyond literal perception, they still focus on inferential VideoQA or inter-frame relation reasoning rather than broader subtext understanding in open online videos.

\paragraph{Humor understanding, meme interpretation, and social meaning.}
A closely related line of work studies higher-level interpretation in humorous or socially contextualized media. v-HUB~\cite{shi2025v} focuses on multimodal video humor understanding, especially in non-verbal short videos, while AVMeme Exam~\cite{jiang2026avmeme} extends evaluation to contextual and cultural understanding of Internet audio-visual memes. Related audio benchmarks, including Dynamic-SUPERB~\cite{huang2024dynamic}, AudioBench~\cite{wang2025audiobench}, MMAU~\cite{sakshi2024mmau}, and MMAR~\cite{ma2025mmar}, mainly evaluate recognition, captioning, dialogue, and semantic or reasoning abilities over audio content.
Closely related humor benchmarks such as FunQA~\cite{xie2024funqa} study surprising or humorous video comprehension, yet are still narrower than the broader space of socially and culturally grounded subtext. In parallel, meme-oriented benchmarks in static image-text settings, including Hateful Memes~\cite{kiela2020hateful}, What Do You Meme?~\cite{sharma2023you}, GOAT-Bench~\cite{lin2024goat}, MemeSafetyBench~\cite{lee2025vision}, and MemeReaCon~\cite{zhao2025memereacon}, probe implicit social meaning, safety, and contextual meme understanding, but cannot capture the temporal, auditory, and evolving multimodal cues that are central to video subtext. In contrast, our focus is on structured, hint-free understanding of video subtext, where models must infer latent meaning from jointly evolving visual, auditory, temporal, and social signals.

\paragraph{Position of ViMU.}
Our work is most closely related to these recent efforts, but differs in both scope and evaluation philosophy. Compared with general video benchmarks, ViMU targets meaning that is not exhausted by visible objects, actions, or temporal relations. Compared with previous works~\cite{chen2025looking,swetha2026vrrqavisualrelationalreasoning}, ViMU is not limited to implicit question answering or hidden inter-frame relations, but instead evaluates whether models can move from observable content to latent subtext, including social signals or culturally grounded interpretations. Compared with humor- or meme-centric benchmarks~\cite{shi2025v,jiang2026avmeme}, ViMU focuses broadly on subtext understanding in videos through a structured taxonomy and hint-free questioning, so that models must recover the intended reading without being given the relevant latent evidence or interpretive hypothesis in advance.

\begin{figure}[t]
    \centering
    \begin{subfigure}[b]{0.49\textwidth}
        \centering
        \includegraphics[width=\textwidth]{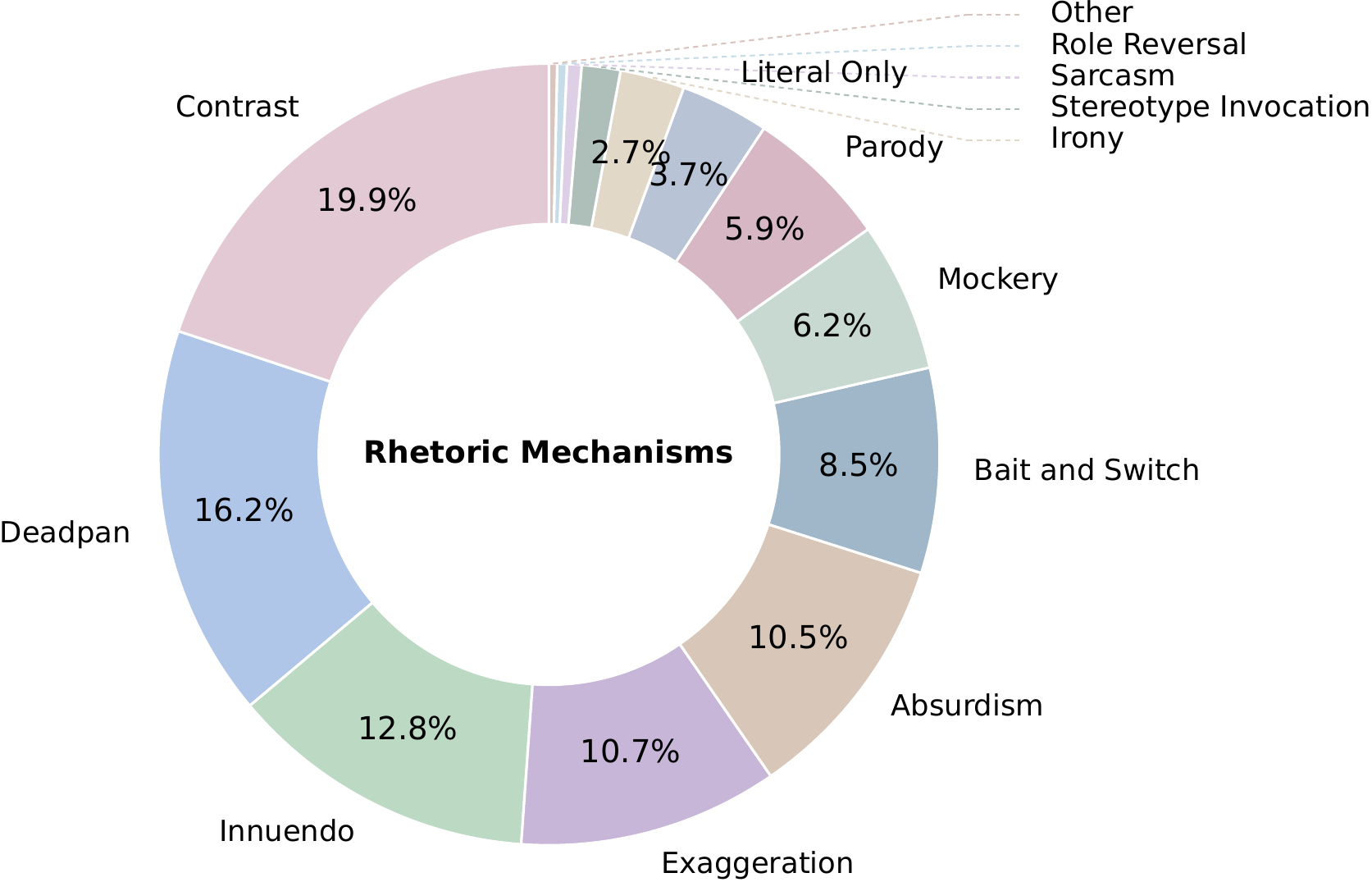}
        \label{fig:rm_donut}
    \end{subfigure} 
    \hfill
    \begin{subfigure}[b]{0.49\textwidth}
        \centering
        \includegraphics[width=\textwidth]{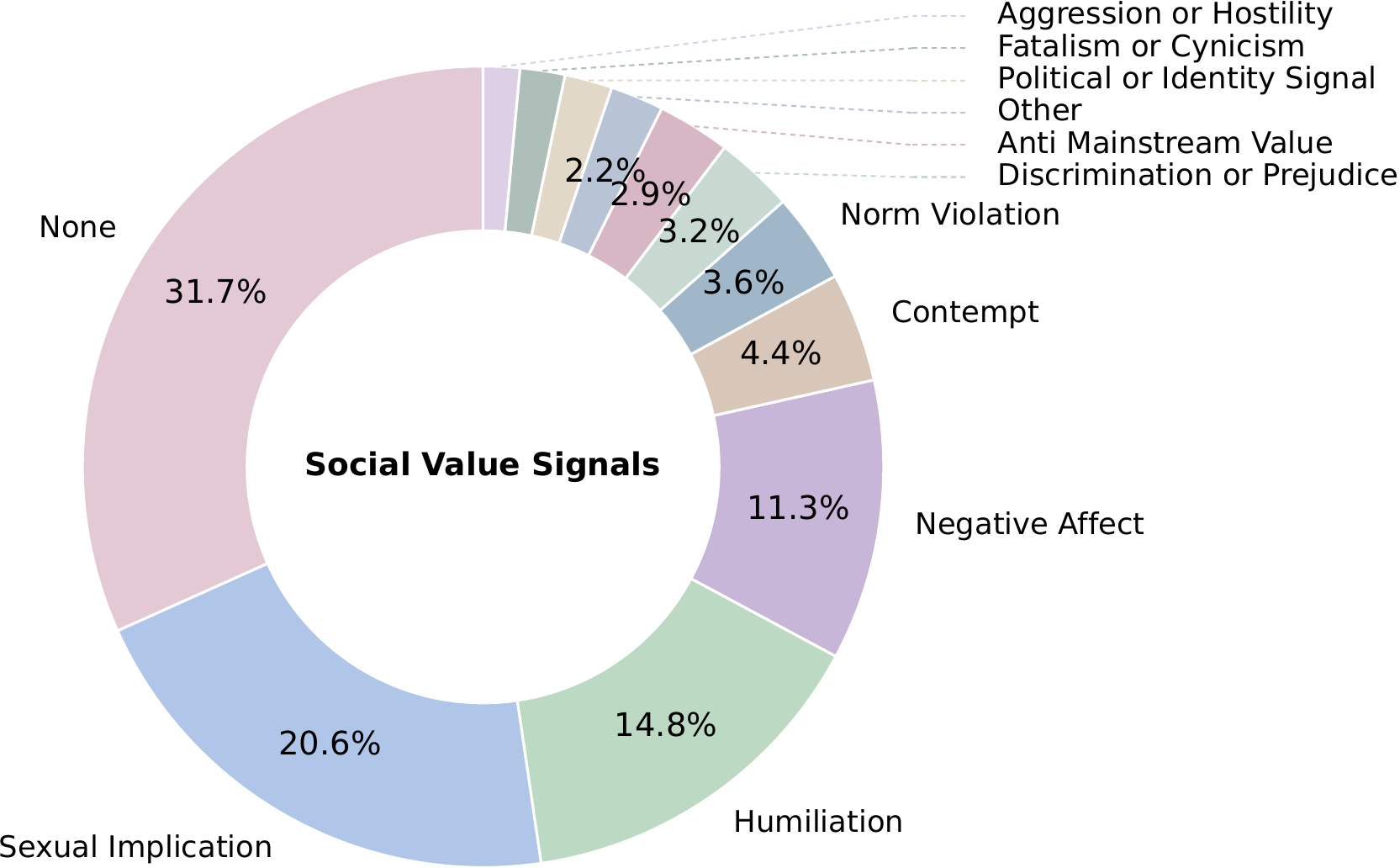}
        \label{fig:sv_donut}
    \end{subfigure} 
    \hfill
    \vspace{-3mm}
    \caption{Distribution of rhetorical mechanisms (left) and social value signals (right) in the dataset.
    The benchmark covers a wide range of rhetorical devices used to construct implicit meaning (left) and the social attitudes or value stances conveyed by videos (right), reflecting diverse forms of non-literal and socially contextualized video communication.
    }
    % \vspace{-2mm}
    \label{fig:donuts}
\end{figure}
\begin{figure}[t]
    \vspace{-5mm}
    \centering
    \begin{subfigure}[c]{0.49\textwidth}
        \centering
        \includegraphics[width=\linewidth,height=4.8cm,keepaspectratio]{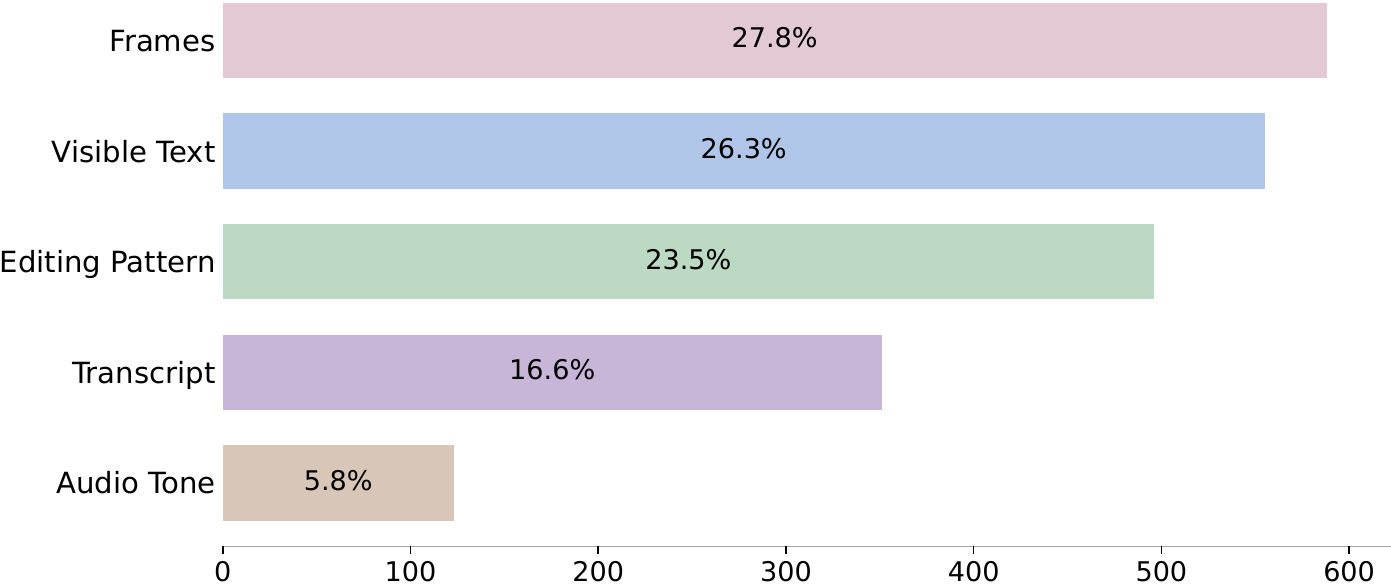}
        \label{fig:es_bar}
    \end{subfigure}
    \hfill
    \begin{subfigure}[c]{0.49\textwidth}
        \centering
        \includegraphics[width=\linewidth,height=4.8cm,keepaspectratio]{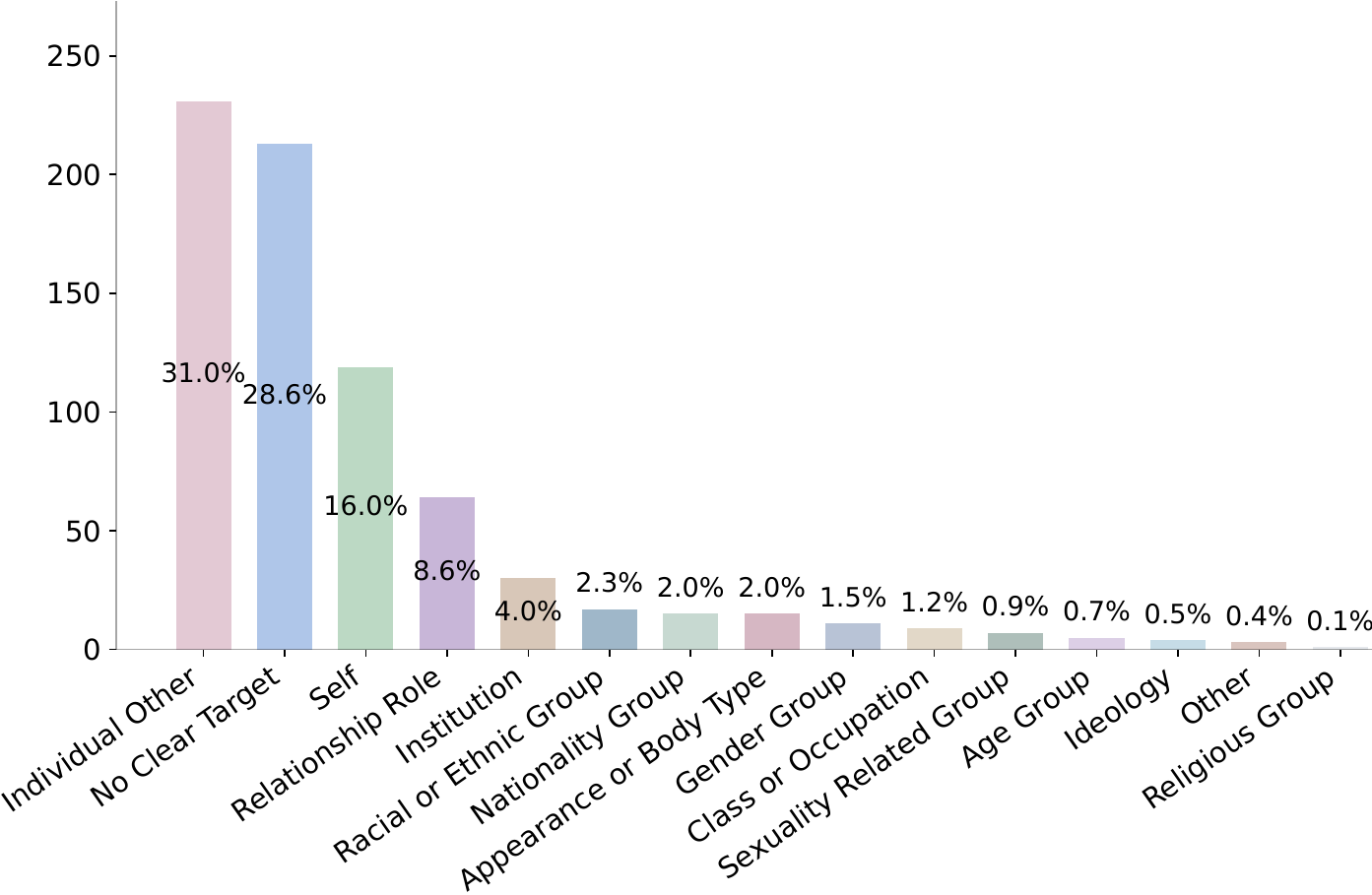}
        \label{fig:ts_bar}
    \end{subfigure}
    \vspace{-5mm}
    \caption{Distribution of evidence sources (left) and target subjects (right) in the dataset.
    The dataset includes multiple types of interpretive evidence, such as visual frames, visible text, editing patterns, transcripts, and audio tone (left), as well as diverse target subjects ranging from individuals and social roles to institutions and identity-related groups (right).
    }
    \vspace{-6mm}
    \label{fig:bars}
\end{figure}

\section{ViMU: Video Metaphorical Understanding Benchmark}

\begin{figure}[ht!]
    \centering
    \begin{subfigure}[t]{\textwidth}
        \centering
        \includegraphics[width=\textwidth]{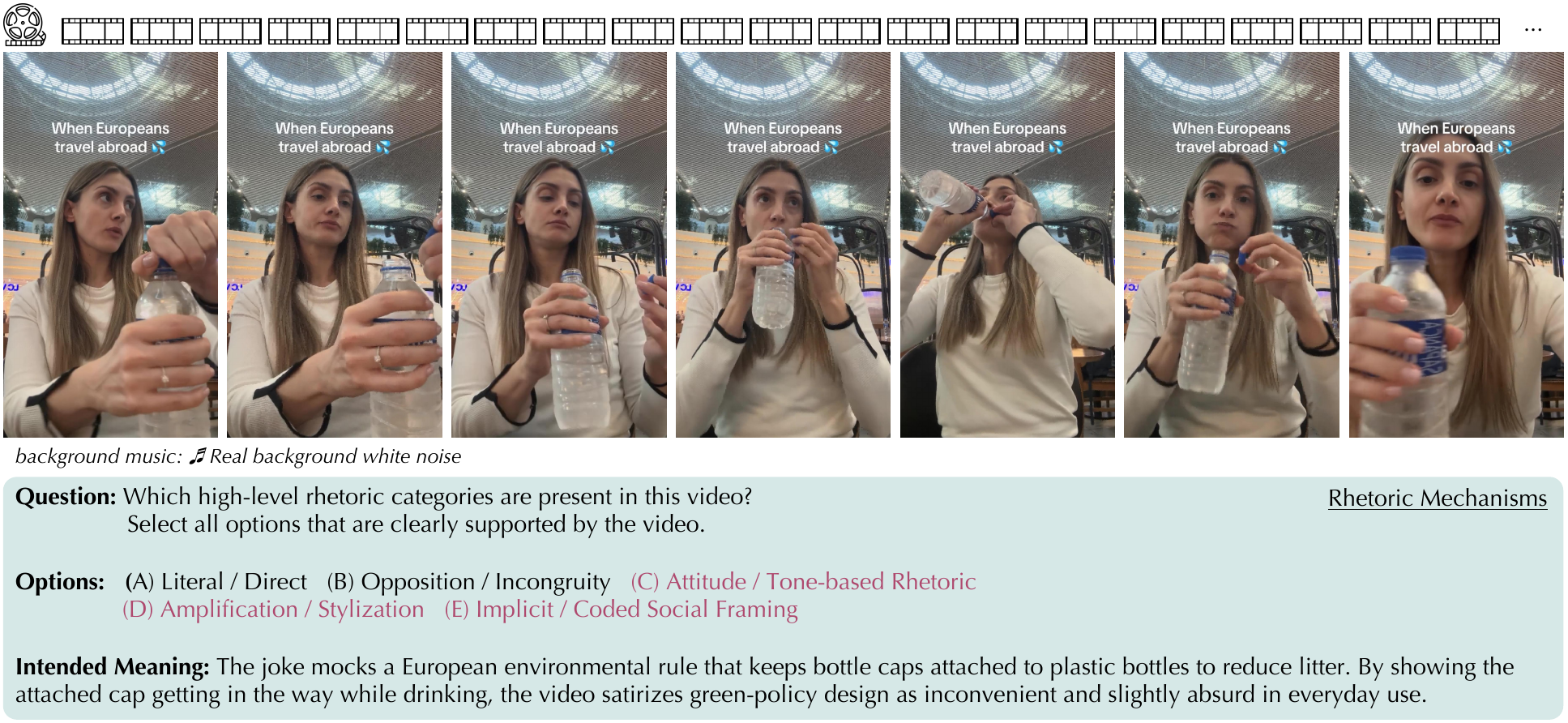}
        \caption{Rhetoric Mechanisms.}
        \label{fig:vimu_example_rhetoric}
    \end{subfigure}
    \begin{subfigure}[t]{\textwidth}
        \centering
        \includegraphics[width=\textwidth]{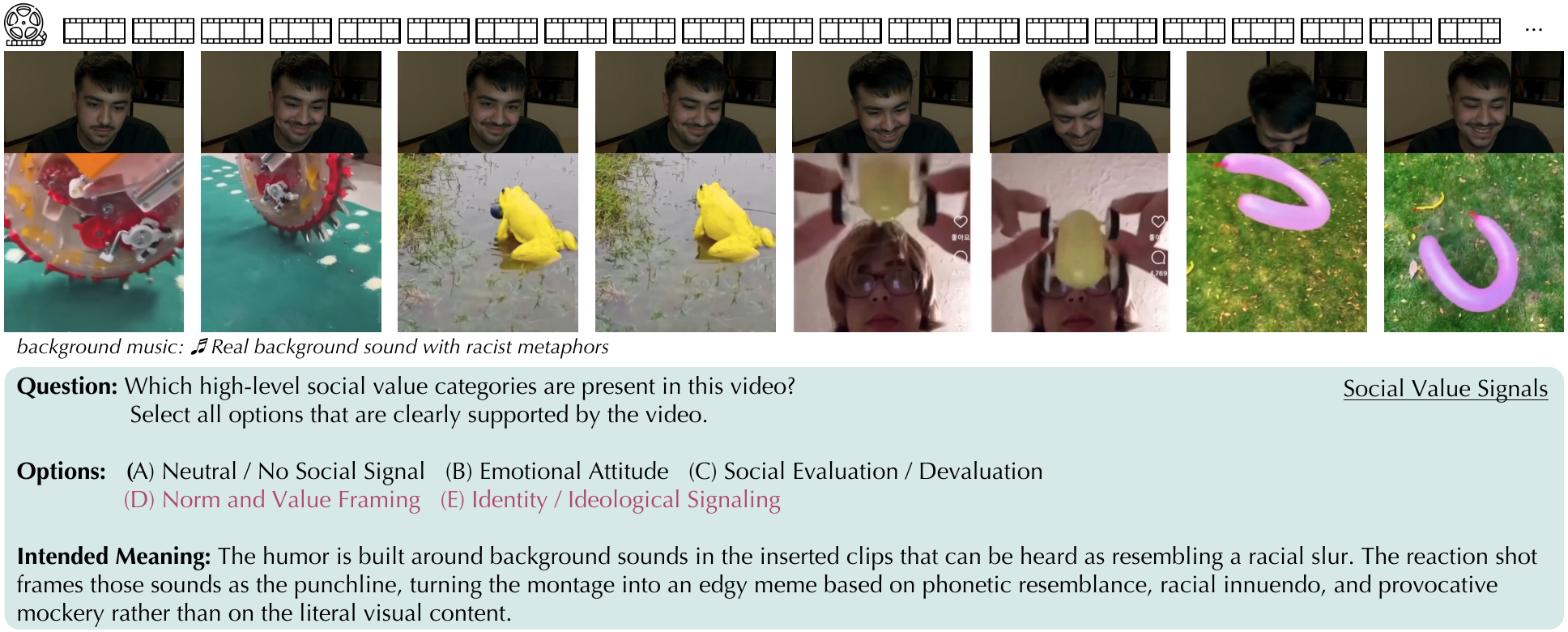}
        \caption{Social Value Signals.}
        \label{fig:vimu_example_social}
    \end{subfigure}
    \begin{subfigure}[t]{\textwidth}
        \centering
        \includegraphics[width=\textwidth]{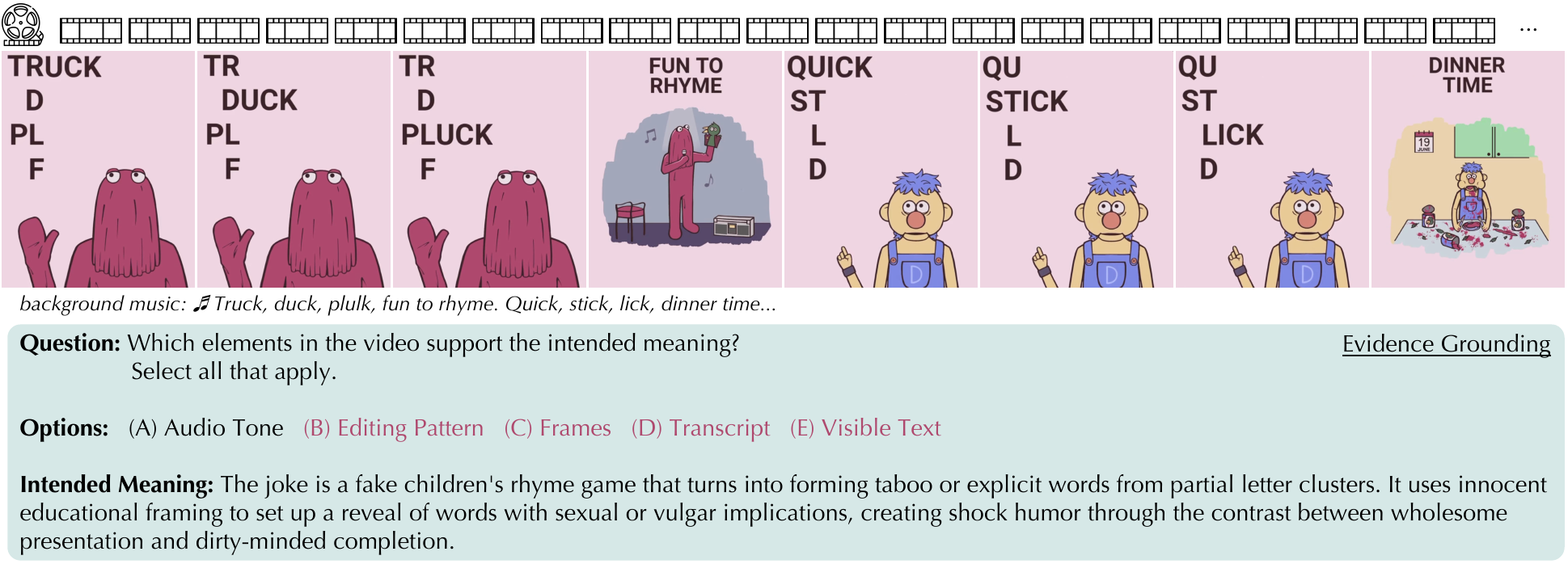}
        \caption{Evidence Grounding.}
        \label{fig:vimu_example_evidence}
    \end{subfigure}
    \caption{Examples of three-types of multiple-choice tasks in ViMU.
    From top to bottom: \textit{evidence grounding}, \textit{rhetoric mechanisms}, and \textit{social value signals}. 
    Each question has five candidate choices, and the ground-truth answers are marked in purple.}
    \label{fig:vimu_mc_examples}
    \vspace{-4mm}
\end{figure}

ViMU is a multi-task benchmark consisting of 2,352 questions from 588 videos across more than ten rhetoric mechanisms and social value signals, specifically designed for video metaphorical understanding, i.e., u derstanding the subtext meaning beyond the surface-level video content. The benchmark is distinguished by the following features.

\begin{figure}[ht!]
    \centering
    \vspace{-3mm}
    \includegraphics[width=\textwidth]{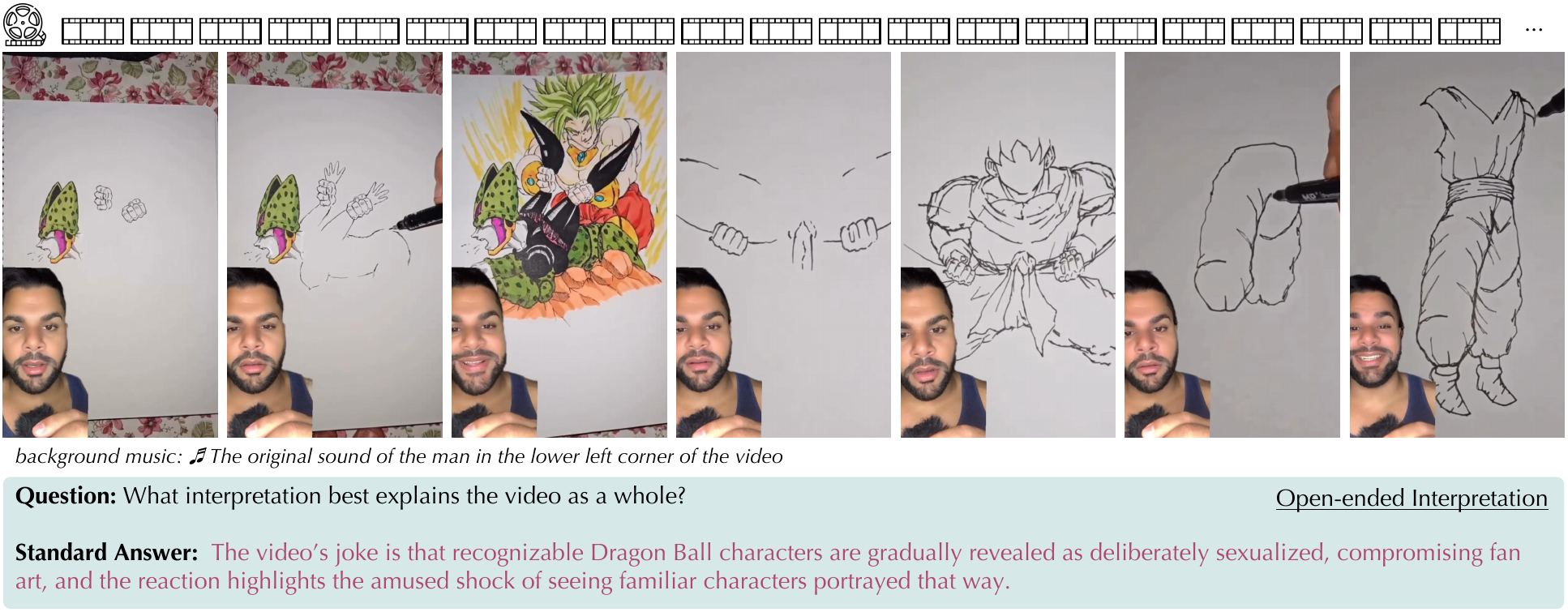}
    \caption{An example of the open-ended interpretation task in ViMU. MLLMs are asked to interpret the video based on the video input, textual prompt, and audio transcript when applicable.}
    \label{fig:vimu_open_example}
    \vspace{-4mm}
\end{figure}
\noindent\textbf{Diversified Semantic Categories.}
As illustrated in Figure~\ref{fig:donuts}, our benchmark spans a diverse set of video categories along two complementary semantic dimensions: \textit{rhetoric mechanisms} and \textit{social value signals}. 
\textit{Rhetoric mechanisms} refer to the communicative devices through which a video conveys its implicit meaning, such as irony, exaggeration, contrast, deadpan delivery, parody, or bait-and-switch. These mechanisms capture how humor, critique, or commentary is constructed at the level of expression.
\textit{Social value signals}, in contrast, describe the underlying social stance, attitude, or normative implication conveyed by the video. These signals capture what the video expresses about social values, emotions, or group relations, including contempt, norm violation, aggression, anti-mainstream sentiment, and others. In shorts, rhetorical mechanisms define how a video should be interpreted, while social value signals capture the stance it conveys. Together, these two dimensions separate \emph{how} meaning is conveyed from \emph{what} social meaning is being expressed. Modeling both enables a more comprehensive evaluation of video metaphor understanding beyond literal perception.

\noindent\textbf{Variety of Evidence Sources and Target Subjects.}
\textit{Evidence sources} refer to the observable cues (e.g., video frames, audios, on-screen text) within a video that support the interpretation of its implicit meaning. The distribution of different evidence sources reflects the multimodal nature of video communication. \textit{Target subjects} describe the entities or groups toward which the video's rhetorical stance or social commentary is directed (e.g., individuals, social groups, institutions, or broader identity categories). Together, these dimensions reveal the wide range of interpretive cues and social referents present in the dataset, supporting comprehensive evaluation of video understanding models.

\noindent\textbf{Comprehensive Evaluation Tasks.}
ViMU provides diversified evaluation tasks to probe complementary aspects. Specifically, the benchmark includes an open-ended interpretation task for evaluating overall understanding of the video’s intended meaning, multi-choice tasks for identifying rhetorical mechanisms and social value signals, and an evidence grounding task for selecting the elements that support the interpretation. Together, these tasks enable a comprehensive evaluation of whether models can understand what a video means, how that meaning is constructed, what social stance it conveys, and whether their interpretations are grounded in observable evidence.

\subsection{Construction of ViMU}
We categorize the tasks into three types according to the level of semantic reasoning required: 
1) \textit{interpretation-level understanding}, which requires inferring the overall intended meaning of the video; 
2) \textit{semantic-structure understanding}, which focuses on identifying the rhetorical mechanisms and social value signals underlying the video; and 
3) \textit{evidence-grounded understanding}, which examines whether models can identify the multimodal evidence supporting their interpretation. 
The construction process of ViMU is discussed with respect to these three categories.

To ensure the task is meaningful and fairly reflects model utility, the dataset construction follows several key principles: (i). Ensuring broad coverage of diverse \textbf{rhetorical mechanisms and social value signals}.
(ii). Given the nature of the task, careful consideration is given to both the sources of \textbf{implicit meaning and the targets of reference}. Implicit cues may arise from visual frames, on-screen text, editing pattern, audio content, or vocal tone. Targets may refer to individuals, other people in the video, or external groups or events not explicitly shown. (iii). For open-ended questions, \textbf{no explicit answer cues are allowed}, as such hints would significantly reduce task difficulty (e.g., directly asking which symbol is being mimicked by the girl through her body movements in Figure~\ref{fig:pipeline} would undermine the task).
Following these principles, we curate over 500 videos from platforms like YouTube, Bilibili, and TikTok, covering more than 10 types of rhetorical mechanisms and social value signals (Figure~\ref{fig:donuts}, detailed explanations of each type are provided in Appendix~\ref{tax_rm} and~\ref{tax_sv}). In addition, as shown in Figure~\ref{fig:bars}, the dataset exhibits strong diversity in evidence sources and target subjects, spanning three modalities (text, vision, audio), five types of evidence sources, and over 10 target categories. This multi-level diversity enables comprehensive evaluation and analysis of model performance. Annotation of these categories and enforcement of hint-free open-ended tasks are achieved through iterative validation by frontier models and human experts. Details are given in Appendix~\ref{app_curation}. Questions regarding different aspects are discussed below.

\begin{table*}[t]
\centering
\caption{\textbf{Main results on ViMU across open-ended interpretation (OE), evidence grounding (EG), rhetoric mechanisms identification (RM), and social value signal identification (SV).}
All values are percentage scores (\%).
\textbf{SSU-Avg} denotes the average of the two structured subtext understanding tasks, RM and SV.
\textbf{All-Avg} denotes the average across all four tasks.
Green shades mark the top-3 models in each metric column, and purple shades mark the bottom-3 models.}
\label{tab:vimu_main_results_release}
\small
\setlength{\tabcolsep}{4.5pt}
\renewcommand{\arraystretch}{1.12}
\resizebox{0.8\textwidth}{!}{
\begin{tabular}{llcccc|c|c}
\toprule
\textbf{Model} & \textbf{Date} & \textbf{OE} & \textbf{EG} & \textbf{RM} & \textbf{SV} & \textbf{SSU-Avg} & \textbf{All-Avg} \\
\midrule
\multicolumn{8}{c}{\textit{Open-weight Models}} \\
\midrule
Ministral-8B           & 2024-10 & \worstB{48.25} & 48.60 & 31.87 & 10.45 & 21.16 & 34.79 \\
Ministral-14B          & 2025-12 & 52.19 & 55.73 & 27.29 & \worstB{6.57} & 16.93 & 35.45 \\
Gemma-3-4B-it          & 2025-03 & \worstA{39.43} & \worstC{25.41} & 21.10 & \worstC{7.17} & 14.13 & \worstC{23.28} \\
Gemma-3-27B-it         & 2025-03 & 55.90 & 49.38 & 32.47 & 7.95 & 20.21 & 36.43 \\
Qwen3-VL-32B-Instruct  & 2025-10 & 64.09 & 59.64 & 27.65 & 15.17 & 21.41 & 41.64 \\
Qwen3.5-27B            & 2026-02 & 62.80 & 60.28 & \bestA{38.18} & 22.40 & 30.29 & \bestC{45.91} \\
\midrule
\multicolumn{8}{c}{\textit{Closed-source / API Models}} \\
\midrule
Claude-3-Haiku         & 2024-03 & 50.41 & 34.55 & \worstB{2.99} & \worstA{3.64} & \worstA{3.32} & \worstB{22.90} \\
GLM-4.5v               & 2025-08 & 62.52 & \worstB{23.11} & 8.87 & 9.26 & 9.06 & 25.94 \\
Grok-4.1-Fast          & 2025-09 & 57.62 & 63.84 & \bestB{34.91} & \bestB{28.73} & \bestA{31.82} & \bestB{46.28} \\
Gemini-3-Flash-Preview & 2025-12 & 62.54 & 52.80 & \bestC{33.63} & \bestC{28.26} & \bestC{30.94} & 44.31 \\
Mimo-V2-Omni           & 2026-03 & 64.07 & 48.94 & 21.04 & 18.52 & 19.78 & 38.14 \\
Seed-2.0-Lite          & 2026-03 & 60.84 & \bestB{66.16} & 18.75 & 16.73 & 17.74 & 40.62 \\
o4-mini                & 2025-04 & \bestC{65.27} & 59.63 & 33.21 & \bestA{29.51} & \bestB{31.36} & \bestA{46.91} \\
GPT-4.1-nano           & 2025-04 & \worstC{50.12} & \worstA{22.31} & \worstA{2.32} & 9.02 & \worstB{5.67} & \worstA{20.94} \\
GPT-5.2                & 2025-12 & \bestA{73.15} & \bestA{67.83} & 16.55 & 21.15 & 18.85 & 44.67 \\
GPT-5.4-mini           & 2026-03 & \bestB{66.19} & \bestC{64.45} & \worstC{4.17} & 11.77 & \worstC{7.97} & 36.64 \\
\bottomrule
\end{tabular}
}
\vspace{-3mm}
\end{table*}

\begin{figure}[t]
    % \vspace{-5mm}
    \centering
    \begin{subfigure}[c]{0.33\textwidth}
        \centering
        \includegraphics[width=\linewidth,height=4.8cm,keepaspectratio]{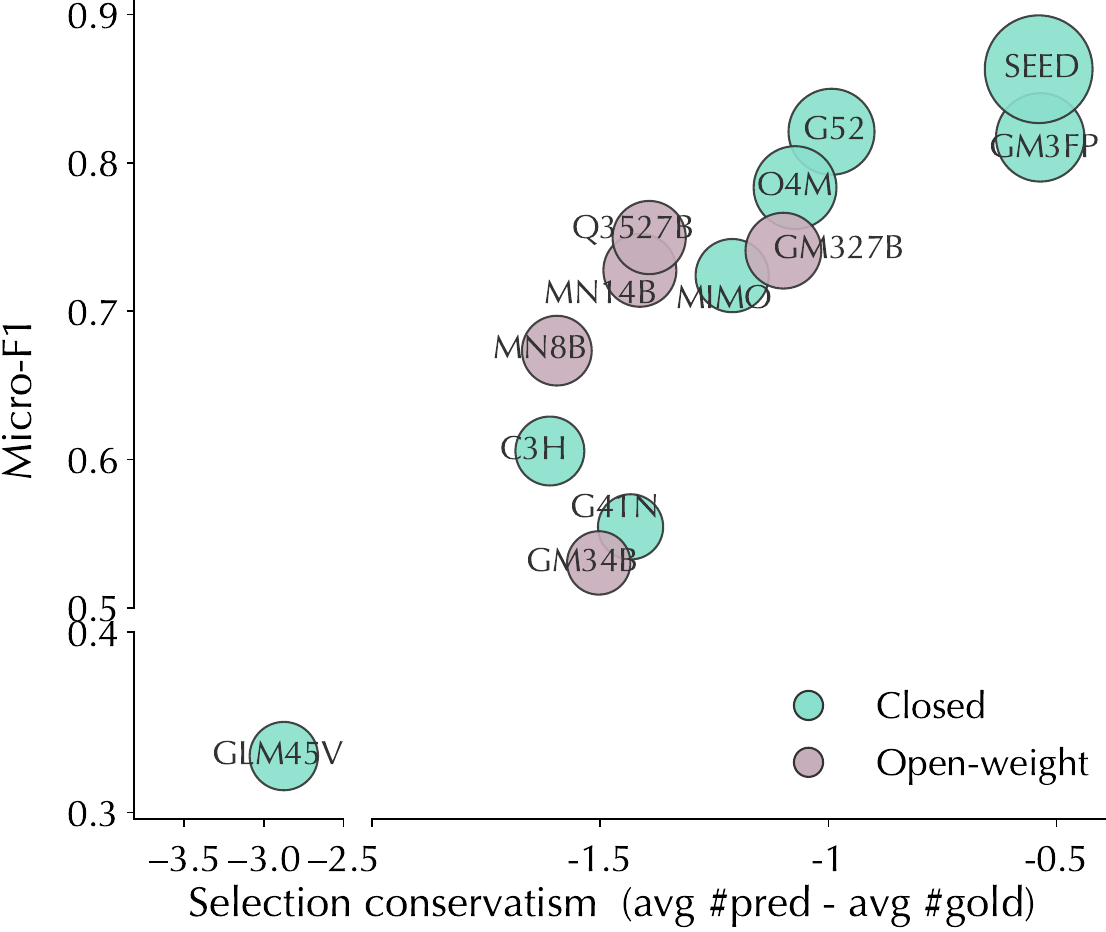}
        \caption{Conservatism vs.\ performance}
        \label{fig:eg1}
    \end{subfigure}
    \hfill
    \begin{subfigure}[c]{0.31\textwidth}
        \centering
        \includegraphics[width=\linewidth,height=4.8cm,keepaspectratio]{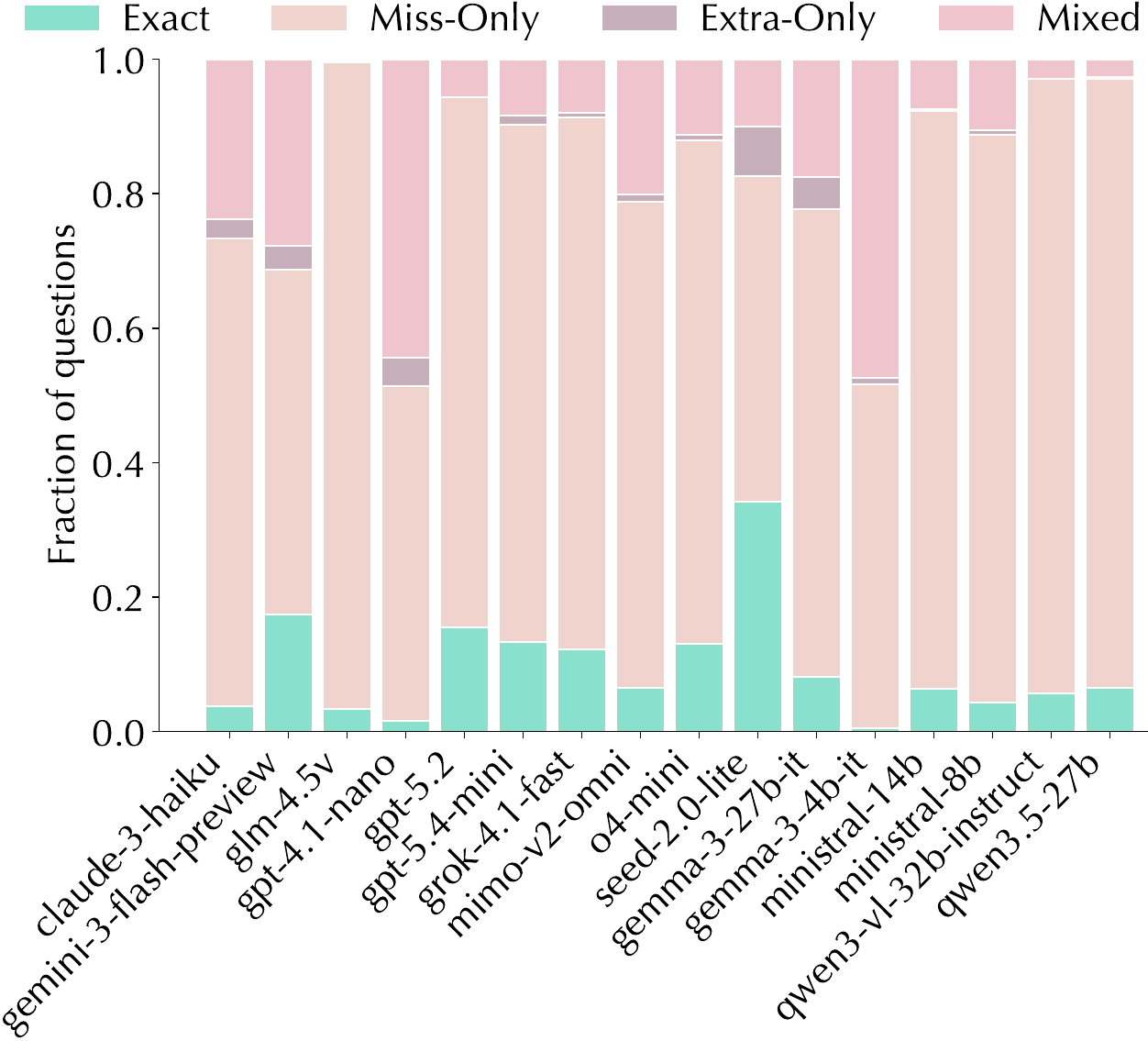}
        \caption{Error composition}
        \label{fig:eg2}
    \end{subfigure}
    \hfill
    \begin{subfigure}[c]{0.34\textwidth}
        \centering
        \includegraphics[width=\linewidth,height=4.8cm,keepaspectratio]{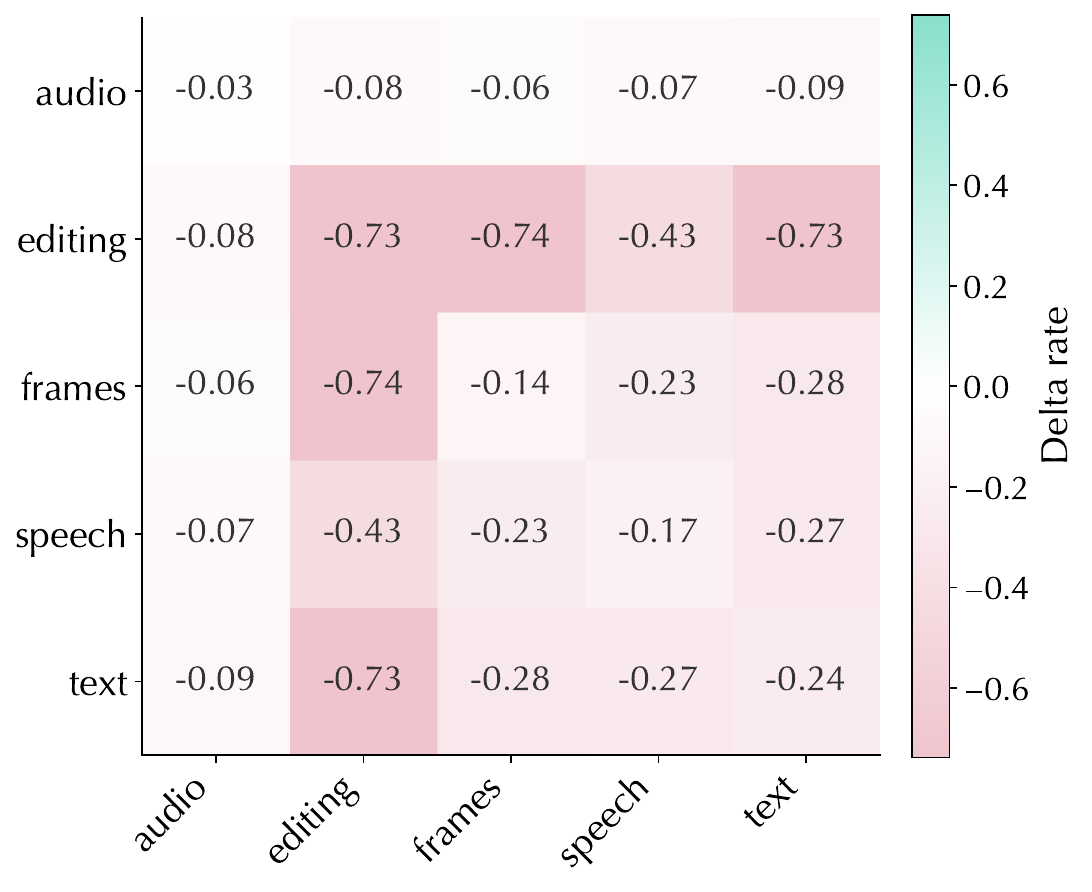}
        \caption{Relation distortion}
        \label{fig:eg3}
    \end{subfigure}
    % \vspace{-3mm}
    \caption{\textbf{Evidence grounding analysis.} From left to right, we show the trade-off between evidence-selection conservatism and grounding quality, the composition of different error types across models, and the overall distortion in pairwise evidence relations relative to the gold co-occurrence structure.}
    \vspace{-5mm}
    \label{fig:eg_ana}
\end{figure}
\subsubsection{Interpretation-Level Understanding}

\noindent\textbf{Open-ended Interpretation (OI).}
This task evaluates whether models can infer the overall meaning conveyed by a video. Given a video clip, the model is asked to explain what the video intends to express as a whole (An example is provided in Figure~\ref{fig:vimu_open_example}). This task requires models to identify the implicit message conveyed through multimodal evidence. The annotation process results in 588 questions. The model responses are evaluated by comparing them with the reference interpretation using a structured grading rubric via LLM-as-a-Judge (details are provided in Appendix~\ref{app_laaj}).

\subsubsection{Semantic-Structure Understanding}

\noindent\textbf{Rhetoric Mechanism Identification (RMI).}
This task requires models to recognize the rhetorical devices used to construct the video's message (An example is provided in Figure~\ref{fig:vimu_example_rhetoric}).
Given a video, the model must select all applicable choices from a predefined list. Here, to improve evaluation
\begin{wrapfigure}{l}{0.48\textwidth}
    \centering
    % \vspace{-0.8em}
    \includegraphics[width=0.48\textwidth]{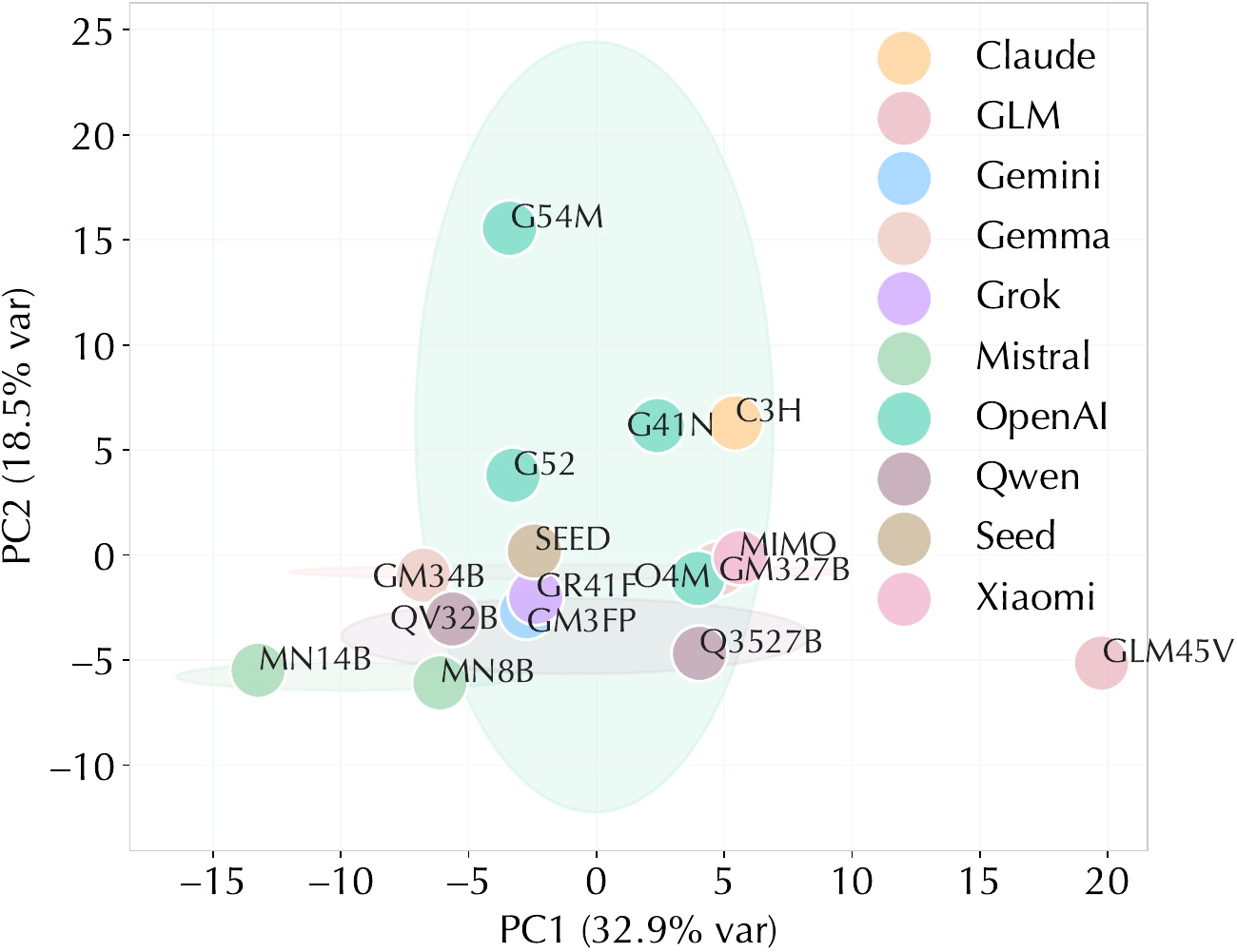}
    \vspace{-0.8em}
    \caption{PCA visualization of model similarity based on error signatures in the macro-5 taxonomy tasks. Each point denotes one model; distances reflect similarity in structured error profiles rather than overall score.}
    \label{fig:model_similarity_pca}
    \vspace{-10mm}
\end{wrapfigure}
stability and interpretability, we further group all rhetorical mechanisms in Figure~\ref{fig:donuts} into five categories (see Appendix~\ref{app_rmgroup} for details). The task is finally formulated as a multiple-choice problem.

\noindent\textbf{Social Value Signal Identification (SVI).}
This task evaluates whether models can identify the social stance or normative implication conveyed by the video (An example is provided in Figure~\ref{fig:vimu_example_social}). Similar to the rhetoric mechanism task, this problem is also formulated as a multiple-choice problem. All the social value signals in Figure~\ref{fig:donuts} are grouped into five categories (Details are provided in Appendix~\ref{app_svgroup}).

\subsubsection{Evidence-Grounded Understanding}

\noindent\textbf{Evidence Grounding (EG).}
This task examines whether models can correctly identify the multimodal evidence supporting their interpretation of the video (An example is provided in Figure~\ref{fig:vimu_example_evidence}). The candidate evidence sources are the five types illustrated in Figure~\ref{fig:donuts}. The task is structured as a multiple-choice problem. This task allows us to analyze whether model reasoning is grounded in observable video cues rather than unsupported speculation.

\section{Experiments and Analysis}

\noindent\textbf{Settings. }We conduct a comprehensive investigation of 16 MLLMs using our ViMU benchmark, encompassing both open-source and proprietary models. For all the considered MLLMs, we employ either a uniform sampling strategy for video processing. All models are evaluated based on their official implementations or available APIs~\cite{openrouter}, with evaluations conducted in a zero-shot manner. More details about the evaluation are provided in Appendix~\ref{app_eval_detial}.

\begin{figure*}[h]
    \centering

    % first row
    \begin{subfigure}[t]{0.32\textwidth}
        \centering
        \includegraphics[width=\linewidth]{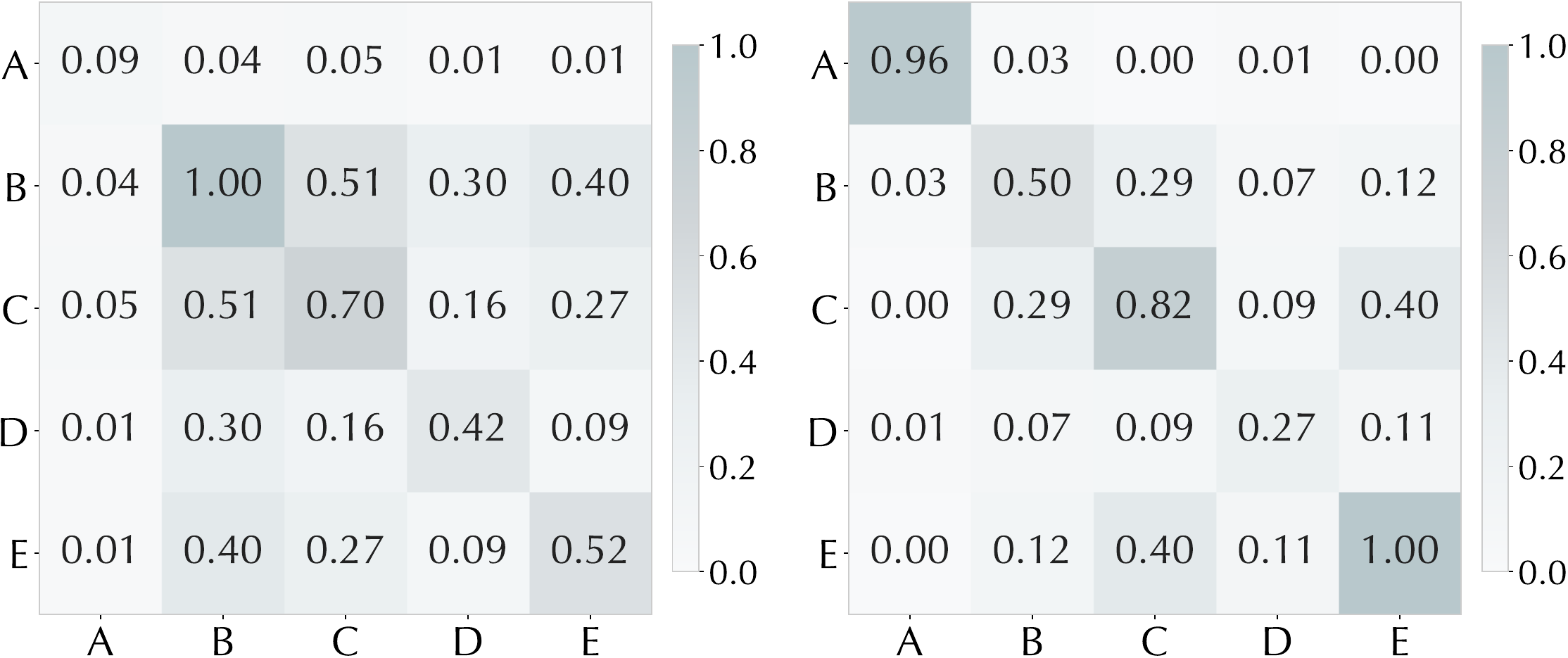}
        \caption{GT co-occurrence structure.}
        \label{fig:tax_geo_a}
    \end{subfigure}
    \hfill
    \begin{subfigure}[t]{0.32\textwidth}
        \centering
        \includegraphics[width=\linewidth]{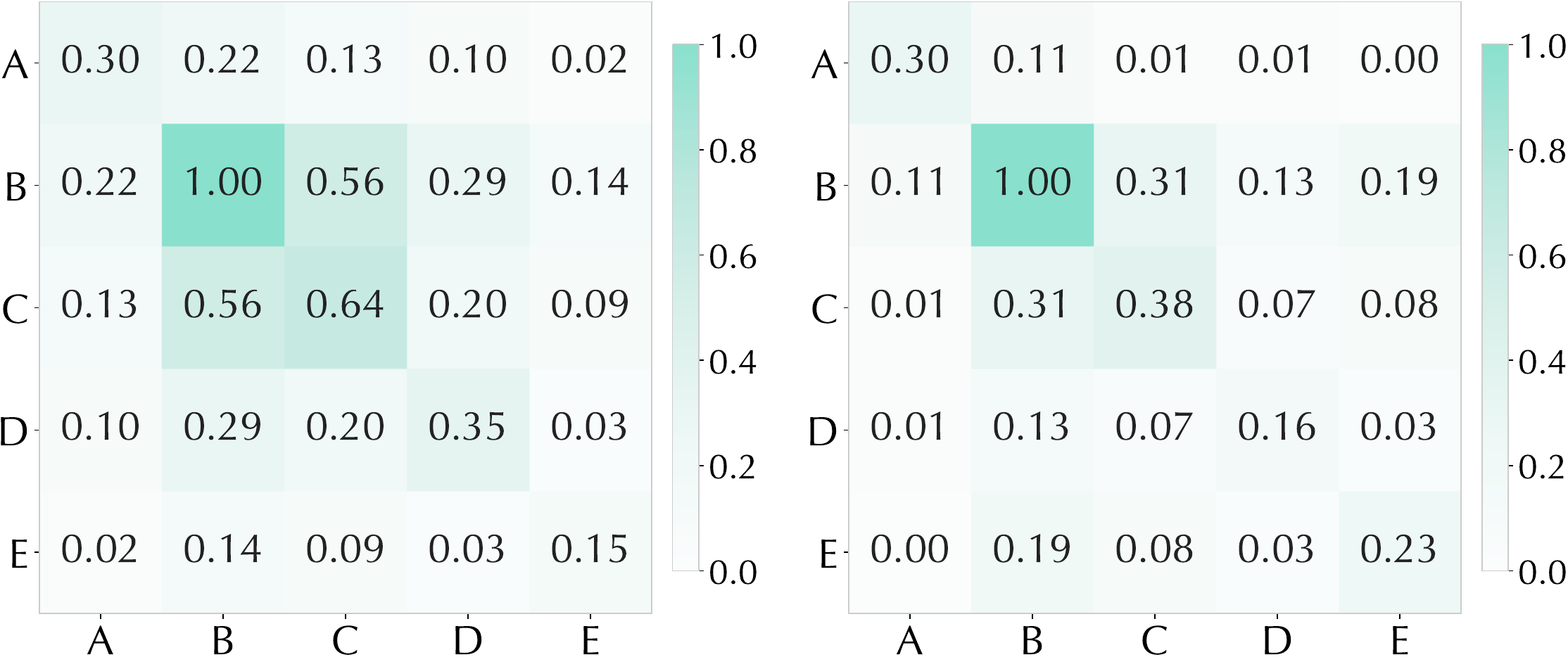}
        \caption{Pr. co-occurrence w.o. guidance.}
        \label{fig:tax_geo_b}
    \end{subfigure}
    \hfill
    \begin{subfigure}[t]{0.33\textwidth}
        \centering
        \includegraphics[width=\linewidth]{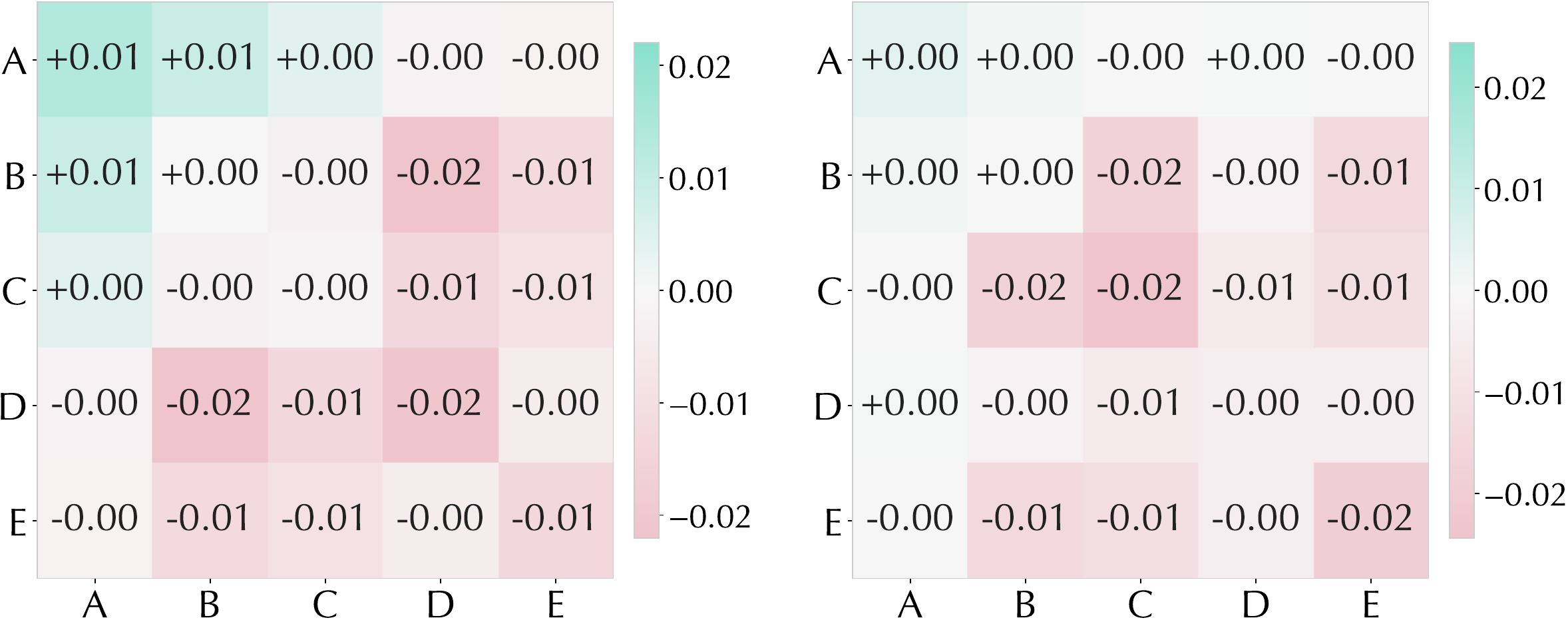}
        \caption{Co-occurrence change.}
        \label{fig:tax_geo_c}
    \end{subfigure}

    \vspace{0.6em}

    % second row
    \begin{subfigure}[t]{0.92\textwidth}
        \centering
        \includegraphics[width=\linewidth]{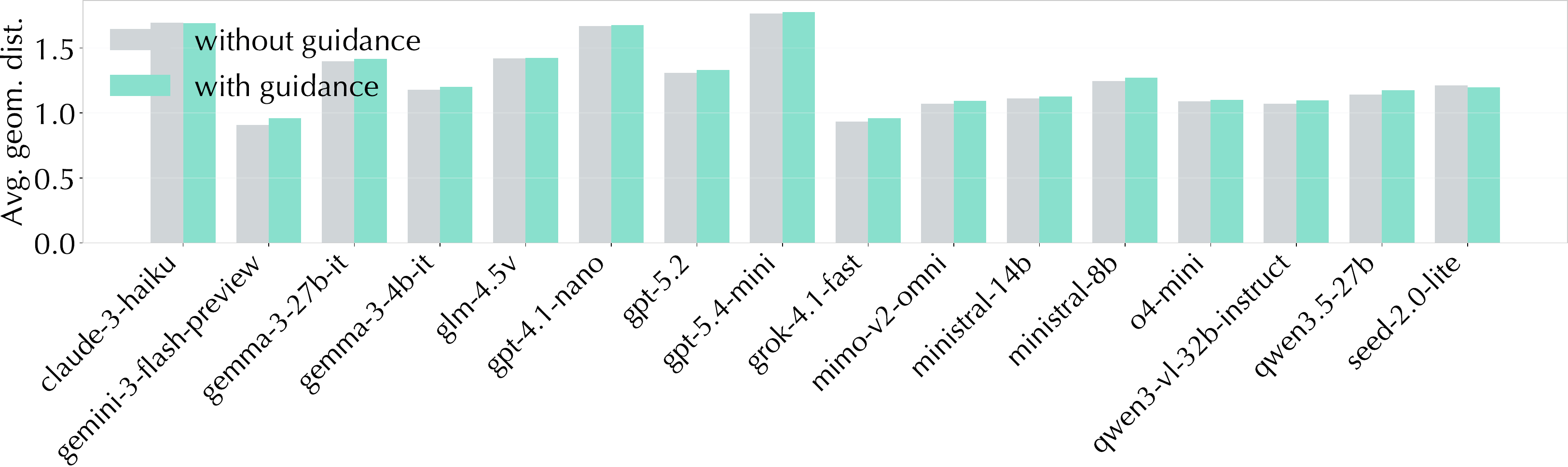}
        \vspace{-6mm}
        \caption{Average taxonomy-geometry distortion across models.}
        \label{fig:tax_geo_d}
    \end{subfigure}

    \caption{Taxonomy geometry analysis of EG and RM predictions. The top row compares the pairwise co-occurrence structure of the ground-truth choices and model predictions. The bottom panel summarizes the average geometry distortion of each model relative to the ground-truth structure.}
    \label{fig:taxonomy_geometry_all}
    \vspace{-2mm}
\end{figure*}
\noindent\textbf{Overall Performance Analysis.} Table~\ref{tab:vimu_main_results_release} reveals a clear pattern: Current models exhibit substantially weaker performance on metaphorical understanding than on general video understanding, which is precisely the gap that ViMU aims to expose. For open-ended interpretation (OE), the strongest performance is achieved by GPT-5.2, which also attains the best evidence grounding (EG) results, both around 70\%. However, when tasked with identifying specific rhetoric mechanisms (RM) and social value signals (SV), its performance drops sharply to around 20\%. In contrast, models such as Grok-4.1-Fast and Gemini-3-Flash-Preview, while less competitive on OE and EG, achieve significantly better results on RM and SV, reaching around 30\%. From these results, we draw three key conclusions: (i) frontier capability in general video interpretation does not automatically translate into precise understanding of implicit stance, rhetorical framing, or socially coded meaning; (ii) different model families, and even models within the same family, exhibit distinct strengths in metaphorical understanding; (iii) closed-source models are not uniformly superior to open-weight models (e.g., Qwen3.5-27B achieves a higher All-Avg than GPT-4.1-nano and Claude-3-Haiku). From a benchmark perspective, these results show that ViMU isolates hidden communicative reasoning and exposes its gap with standard video understanding.

\noindent\textbf{Analysis on Evidence Grounding (EG).}
Figure~\ref{fig:eg1} visualizes how each model trades off evidence-selection conservatism against overall grounding quality: the x-axis measures whether a model tends to under-select or over-select evidence relative to the gold answer, while the y-axis reports its Micro-F1. For readability, we abbreviate model names as follows:
\textbf{C3H} = \texttt{claude-3-haiku},
\textbf{GM3FP} = \texttt{gemini-3-flash-preview},
\textbf{GLM45V} = \texttt{glm-4.5v},
\textbf{G41N} = \texttt{gpt-4.1-nano},
\textbf{G52} = \texttt{gpt-5.2},
\textbf{MIMO} = \texttt{mimo-v2-omni},
\textbf{O4M} = \texttt{o4-mini},
\textbf{SEED} = \texttt{seed-2.0-lite},
\textbf{GM327B} = \texttt{gemma-3-27b-it},
\textbf{GM34B} = \texttt{gemma-3-4b-it},
\textbf{MN14B} = \texttt{ministral-14b},
\textbf{MN8B} = \texttt{ministral-8b}, and
\textbf{Q3527B} = \texttt{qwen3.5-27b}. Figure~\ref{fig:eg1} therefore characterizes the \emph{selection style} of different models rather than only their final score. As shown, most models lie on the conservative side, indicating that they tend to predict fewer evidence sources than the annotations require (x-axis < 0). Mild conservatism does not necessarily reduce performance, but excessive under-selection is clearly harmful: the most conservative outlier is also among the weakest performers. At the same time, the top closed models occupy the upper region of the figure,  whereas the strongest open-weight models are competitive but still generally fall slightly below the best closed models. Overall, Fig.~\ref{fig:eg1} suggests that the main risk in current evidence grounding models is \textit{not aggressive over-selection, but incomplete retrieval of supporting evidence.}

Figure~\ref{fig:eg2} further decompose each model's predictions into four error types---\textit{Exact}, \textit{Miss-Only},
\begin{wrapfigure}{r}{0.50\textwidth}
    \centering
    \begin{subfigure}[t]{0.48\textwidth}
        \centering
        \includegraphics[width=\linewidth]{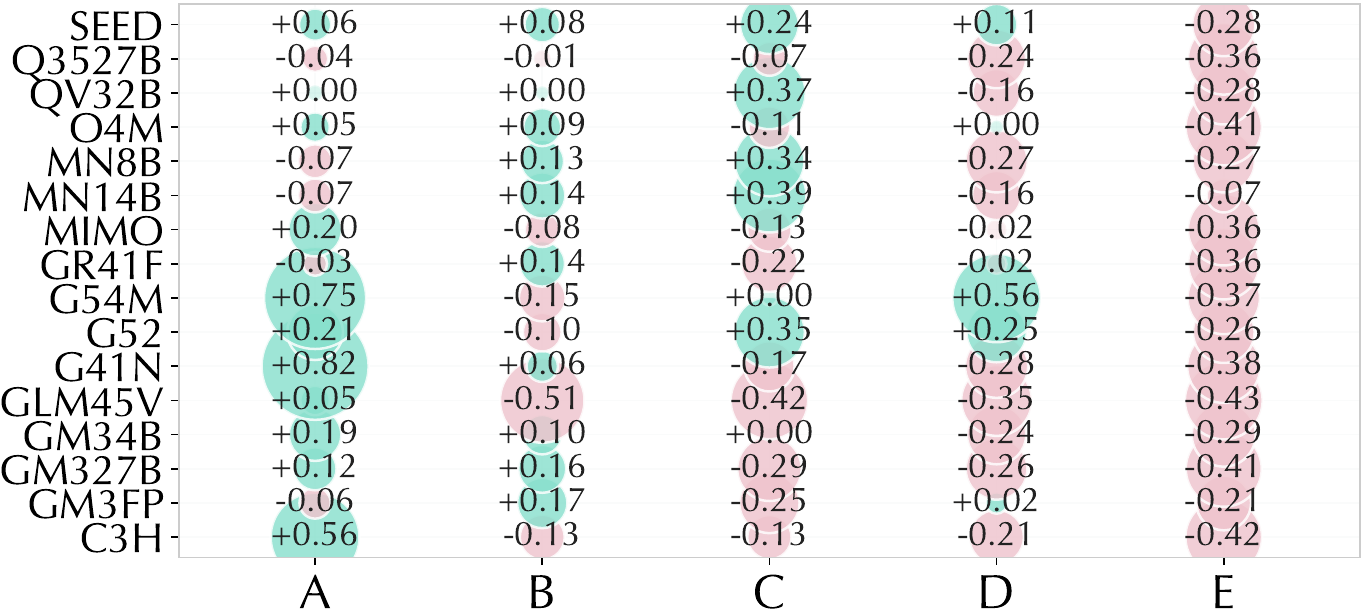}
        \caption{Rhetoric Mechanisms Identification.}
        \label{fig:affinity_rhetoric}
    \end{subfigure}
    \begin{subfigure}[t]{0.48\textwidth}
        \centering
        \includegraphics[width=\linewidth]{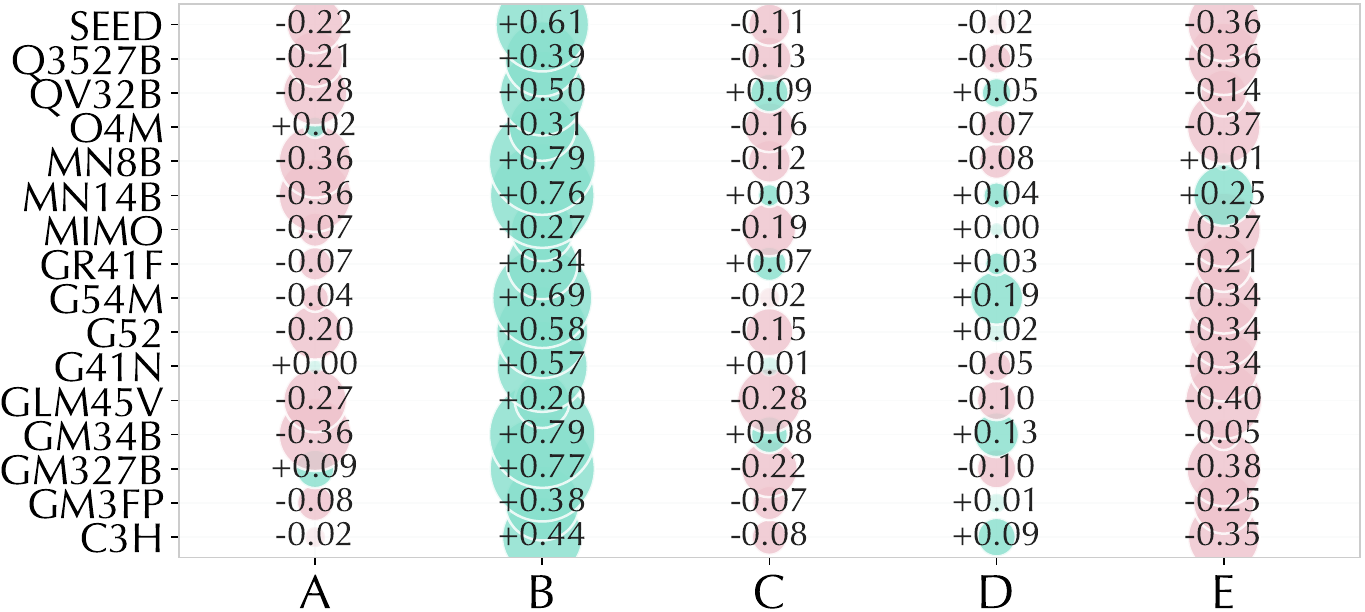}
        \caption{Social Value Signal Identification.}
        \label{fig:affinity_social}
    \end{subfigure}
    \vspace{-2mm}
    \caption{Model--option affinity bias without guidance. Positive values indicate over-prediction relative to ground-truth prevalence, while negative values indicate under-prediction.}
    \label{fig:affinity_maps_main}
    \vspace{-5mm}
\end{wrapfigure}
\textit{Extra-Only}, and \textit{Mixed}---so as to show \emph{how} models fail rather than merely \emph{how often} they fail. The figure reveals that a substantial portion of non-exact predictions is driven by omission-related errors, especially \textit{Miss-Only} and \textit{Mixed}, while purely over-selective behavior (\textit{Extra-Only}) is generally less dominant. This confirms the trend already suggested by Fig.~\ref{fig:eg1}: evidence grounding errors are driven more by failing to retrieve all required cues than by indiscriminately hallucinating additional evidence. 

Figure~\ref{fig:eg3} analyzes evidence grounding at the level of \emph{pairwise evidence relations}. Specifically, it compares the average co-selection pattern produced by models against the gold co-occurrence pattern, thereby revealing whether models over-link or under-link different evidence types. The matrix is uniformly negative, which means that, on average, \textit{models under-predict evidence co-occurrence rather than over-connecting evidence sources.} The largest negative deviations involve \textit{editing}-related pairs, especially editing--frames and editing--text, whereas audio-related relations remain much closer to zero. This suggests that current models are relatively better at handling isolated perceptual cues, but substantially weaker at recovering structured multi-source evidence patterns, particularly when editing signals must be integrated with visual or textual evidence.

\noindent\textbf{Analysis on Rhetoric Mechanisms (RM) and Social Value Signals (SV).} Figure~\ref{fig:model_similarity_pca} visualizes model similarity by applying PCA to each model’s error-signature vector on the RM and SV tasks. PC1
\begin{wrapfigure}{l}{0.45\textwidth}
    \centering
    \vspace{-1em}
    \includegraphics[width=0.45\textwidth]{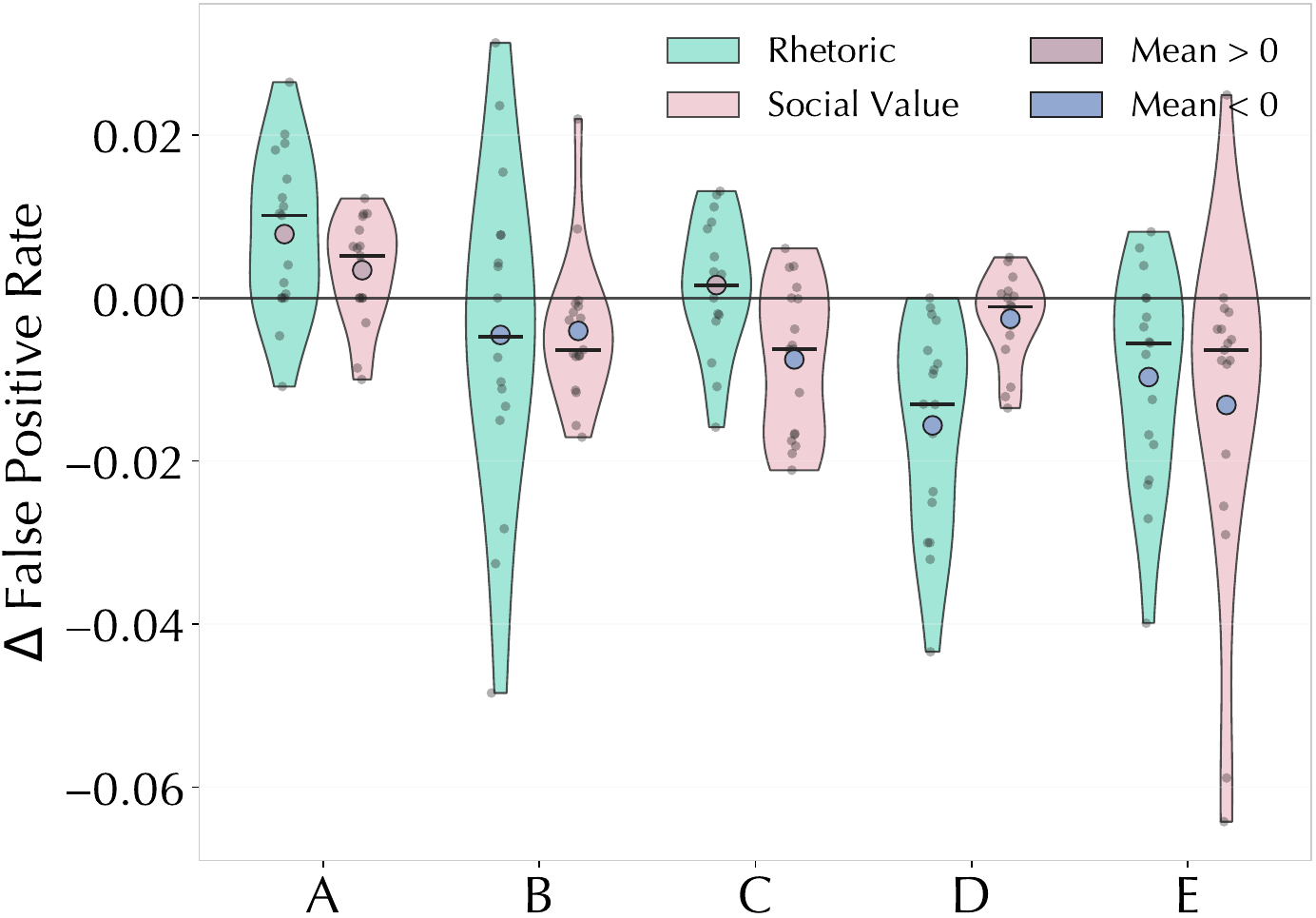}
    \vspace{-0.8em}
    \caption{Category-wise distribution of guidance-induced shifts in false positive rate ($\Delta$FPR). Each violin summarizes the distribution over models for a given category, with rhetoric (green) and social value (red) shown side by side. Points denote model-level values, while markers indicate mean shifts.}
    \label{fig:violin}
    \vspace{-5mm}
\end{wrapfigure}
and PC2 denote the first two principal components, explaining 32.9\% and 18.5\% of the variance, respectively. Note that models lie closer have more \textit{similar structured error profiles}. We observe family-level clustering: OpenAI models tend to be grouped together, and Qwen and Mistral models also remain close to their family peers, indicating shared inductive biases in how they organize taxonomy labels. Overall,Figure~\ref{fig:model_similarity_pca} shows that models with similar overall performance may still differ substantially in their \textit{decision patterns}, and that behavior is strongly shaped by model family.

In Figure~\ref{fig:taxonomy_geometry_all}, we study taxonomy geometry preservation by comparing the pairwise co-occurrence structure of the five ground-truth choices and model predictions. For each task (rhetoric and social value), we construct a normalized (5$\times$5) co-occurrence matrix, where diagonal entries reflect label prevalence and off-diagonal entries capture label interactions. Figure~\ref{fig:tax_geo_a} shows the ground-truth geometry, Figure~\ref{fig:tax_geo_b} shows the average prediction geometry without extra guidance, Figure~\ref{fig:tax_geo_c} shows the change induced by guidance, and Figure~\ref{fig:tax_geo_d} summarizes model-wise distortion using the Frobenius distance to the ground-truth matrix. The guidance information can be found in Appendix~\ref{app_guided_prompts}. As can be observed, models recover part of the taxonomy structure, as several dominant co-occurrence patterns in the annotations also appear in predictions, but the predicted matrices are generally flatter and less contrasted, suggesting that fine-grained relations are only partially preserved. Furthermore, the guidance introduces mostly small but systematic local shifts in pairwise relations, yet these changes \textit{do not consistently improve global structural fidelity}: for many models, the distance to the ground-truth geometry remains similar or becomes slightly larger. Overall, Figure~\ref{fig:taxonomy_geometry_all} indicates that models capture limited taxonomy structure, and guidance mainly reweights local decisions rather than restoring global structure.

The affinity-bias maps in Figure~\ref{fig:affinity_maps_main} reveal clear option-level biases rather than uniform error. In rhetoric, many models over-predict \textit{A (Literal / Direct)} and under-predict \textit{E (Implicit / Coded Social Framing)}, suggesting a tendency to map difficult cases to safer or more generic categories. As for social signal, most models strongly over-predict \textit{B (Emotional Attitude)} while under-predicting \textit{E (Identity / Ideological Signaling)}, indicating that broad affective readings often act as a default interpretation.
We also find that the with-guidance results are qualitatively very similar to the without-guidance ones (See Appendix~\ref{app_wguidance_fig9}), which suggests that guidance does not substantially change the underlying option-allocation structure but only makes small local adjustments.

In Figure~\ref{fig:violin}, we explore how guidance affects model false positive behavior across different categories. The average results of all considered models are report. For rhetoric, categories such as \textit{B (Opposition Incongruity)} exhibit larger variance, suggesting increased instability when models handle contrastive or unexpected structures, while \textit{D (Amplification Stylization)} tends to shift negatively, reflecting more conservative predictions. For social value signals, the overall shifts are more compact but show stronger polarization in categories like \textit{E (Identity Ideological Signaling)}, where models become more conservative yet less consistent across instances. 

\section{Conclusion}
In this work, we introduce \texttt{ViMU}, a benchmark designed to evaluate video models beyond literal perception by focusing on subtext understanding, including rhetorical, social, and culturally grounded meanings. Our results show that, despite strong performance on surface-level tasks, current frontier models struggle substantially with interpreting implicit meaning, achieving below 50\% overall performance. Through fine-grained analyses, we further reveal systematic gaps and distinct behavioral patterns across models. These findings highlight a fundamental limitation of existing video understanding systems and suggest that advancing toward robust, human-like interpretation requires modeling \textbf{not only what is shown, but also what is meant.}

\bibliographystyle{plain}
\bibliography{main}

\newpage

\appendix

\section{Details of the Dataset Curation Process}
\label{app_curation}

\begin{figure}[ht!]
    \centering
    \includegraphics[width=0.9\textwidth]{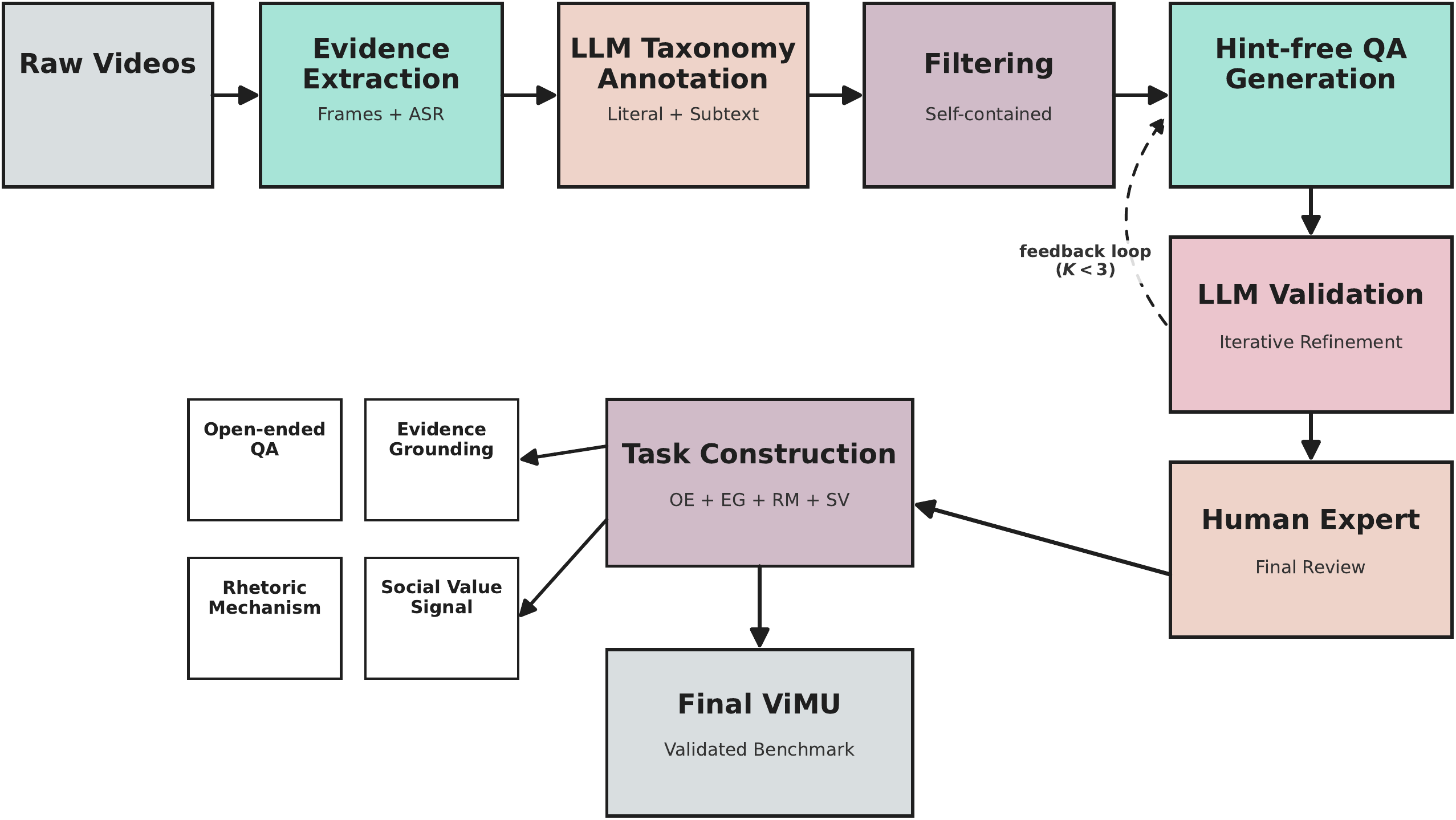}
    \caption{An illustration of the dataset curation process.}
    \label{fig:curation}
\end{figure}

The construction of ViMU follows a multi-stage pipeline that integrates multimodal evidence extraction with LLM-driven semantic annotation and question refinement, as well as human expert review. The overall goal is to produce high-quality, hint-free benchmark instances that require genuine subtext understanding. An illustration of the curation pipeline can be found in Figure~\ref{fig:curation}. 

\noindent\textbf{Stage 1: Multimodal Evidence Extraction.}
Given a set of videos $\mathcal{V} = \{v_i\}$, we construct for each video a multimodal evidence representation by extracting uniformly sampled frames $\mathcal{F}_i$ and an audio transcript $t_i$, yielding:
\[
\mathcal{E}_i = \{\mathcal{F}_i, t_i\}.
\]
This ensures that all downstream reasoning is grounded in observable signals rather than external metadata. 

\noindent\textbf{Stage 2: LLM-based Taxonomy Annotation.}
We employ a frontier modelc (GPT-5.4) to produce structured semantic annotations for each video. The model is prompted to separate literal content from intended meaning and to decompose subtext into multiple axes, including rhetorical mechanisms and social value signals:
\[
\mathcal{T}_i = f_{\text{LLM}}(\mathcal{E}_i).
\]
Annotations rely only on evidence in $\mathcal{E}_i$, ensuring grounding and interpretability.

\noindent\textbf{Stage 3: LLM-based Hint-Free Question Generation.}
Conditioned on the taxonomy $\mathcal{T}_i$, we use the same LLM to generate a question–answer pair $(q_i, a_i)$:
\[
(q_i, a_i) = g_{\text{LLM}}(\mathcal{T}_i).
\]
A key constraint is that $q_i$ must be \emph{hint-free}, i.e., it must not explicitly reveal the semantic dimension (e.g., rhetoric or social meaning) being tested. This prevents shortcut learning and forces genuine inference of the intended meaning. This is achieved through the explicit enforcement in the prompt and the iterative refinement in Stage 4.

\noindent\textbf{Stage 4: Iterative LLM-based Validation and Refinement.}
To ensure quality, each generated QA instance undergoes an iterative refinement loop. At iteration $k$, the LLM evaluates the current QA instance $\mathcal{Q}_i^{(k)}$ and produces structured feedback:
\[
\text{feedback}^{(k)} = h_{\text{LLM}}(\mathcal{T}_i, \mathcal{Q}_i^{(k)}).
\]
The QA is then updated:
\[
\mathcal{Q}_i^{(k+1)} = g_{\text{LLM}}(\mathcal{T}_i, \text{feedback}^{(k)}),
\]
until it satisfies criteria such as implicitness, difficulty, and alignment with intended meaning. This loop explicitly enforces that the question cannot be solved using surface-level cues alone. We allow at most $K=3$ refinement rounds. At round $k=0$, the initially generated QA pair is validated. If it is marked as pass or minor revision, it is accepted; if it is marked as reject, it is discarded. If it receives a major revision label, the feedback is used to regenerate the QA pair for the next round. Samples that still fail after $K$ refinement rounds are rejected.

\noindent\textbf{Stage 5: Evidence and Task Construction.}
Using the annotated evidence fields, we derive a unified set of evidence sources and construct structured evaluation tasks, including evidence grounding and taxonomy classification. Fine-grained labels are further aggregated into macro-level categories (5 categories) to support analysis at different abstraction levels. Details are given in Appendix~\ref{app_rmgroup} and~\ref{app_svgroup}

\noindent\textbf{Filtering and Quality Control.}
We retain only samples that are self-contained and suitable for fair evaluation, filtering out videos that require strong external context or exhibit ambiguous semantics. The dataset is finally validated by 5 human experts. This ensures that performance reflects intrinsic video understanding rather than external knowledge retrieval. The curation process results in a high quality dataset containing 588 videos.

\noindent\textbf{Summary.}
The final dataset is constructed through a pipeline:
\[
v_i \rightarrow \mathcal{E}_i \rightarrow \mathcal{T}_i \rightarrow (q_i, a_i) \rightarrow \mathcal{Q}_i^*,
\]
where $\mathcal{Q}_i^*$ denotes the validated QA instance. Notably, LLMs are used not only as annotators but also as generators and iterative refiners, enabling scalable yet high-quality benchmark construction.

\definecolor{promptbg}{HTML}{F6F8F8}
\definecolor{promptframe}{HTML}{D0D5D8}

\newtcolorbox{promptbox}{
  colback=promptbg,
  colframe=promptframe,
  boxrule=0.4pt,
  arc=2mm,
  left=1mm,
  right=1mm,
  top=1mm,
  bottom=1mm,
  fontupper=\small,
  breakable
}

\subsection{Prompt Design Details}
\label{app_prompts}

In this section, we summarize the key prompts used for taxonomy annotation, question generation, and iterative validation as follows.

\begin{tcolorbox}[promptbox,title={Taxonomy Annotation Prompt}]
Video ID: \texttt{\textless video\_id\textgreater}

Duration (sec): \texttt{\textless duration\textgreater}

\textbf{Task:}
Analyze this video meme in a hierarchical, multi-axis way.

\textbf{Available evidence:}

1. Video frames

2. Transcript from audio ASR (may be empty, noisy, or partial):

\texttt{\textless transcript\textgreater}

\textbf{Important constraints:}

- You must infer visible on-screen text directly from the frames when relevant.

- Do not assume any external posting context beyond the video itself.

- Separate \textbf{literal content} from \textbf{intended meaning}.

- Use only evidence supported by \texttt{frames}, \texttt{transcript}, \texttt{visible text}, \texttt{audio tone}, or \texttt{editing/timing cues}.

- If the video is not self-contained, reflect that in the \texttt{interpretability bucket}.
\end{tcolorbox}

This prompt corresponds exactly to the user-level input used for taxonomy labeling.

\begin{tcolorbox}[promptbox,title={Hint-Free Question Generation Prompt}]
Video ID: \texttt{\textless vid\_id\textgreater}

You are given the annotation for one video meme.

\textbf{Taxonomy JSON:}

\texttt{\textless taxonomy json\textgreater}

\textbf{Task:}

Create exactly one difficult, hint-free question for evaluating a video understanding model.

\textbf{Hard constraints:}

- The question must NOT contain words that leak the target semantic field, such as:
  \texttt{hate}, \texttt{bias}, \texttt{stereotype}, \texttt{discrimination}, \texttt{offensive}, \texttt{toxic}, \texttt{negative}, \texttt{sarcasm}, \texttt{irony}, \texttt{prejudice}, \texttt{controversial}

- The question should not presuppose that the video has hidden meaning.

- The question should require understanding the intended meaning, not just literal description.

- The gold answer should be concise but semantically complete.

- The rubric should make later LLM judging reliable.
\end{tcolorbox}

This prompt is used for initial QA generation.

\begin{tcolorbox}[promptbox,title={Validation Prompt}]
\textbf{Taxonomy JSON:}

\texttt{\textless taxonomy json\textgreater}

\textbf{Question package JSON:}
\texttt{\textless qa json\textgreater}

\textbf{Validation task:}

Check whether this benchmark question is genuinely hint-free, difficult, and aligned with the intended meaning.

\textbf{Evaluate the following:}

1. Does the question leak the semantic field or sensitive framing?

2. Can the question be answered correctly using only surface-level description?

3. Does it force understanding of the intended meaning?

4. Is the difficulty appropriate given the video's taxonomy?

5. Is the gold answer aligned with the intended meaning?

6. Is the rubric strong enough for later LLM judging?

If the question is flawed, provide a better \texttt{revised\_question} that is harder and less revealing while still evaluable.
\end{tcolorbox}

This prompt defines the structured validation step. 

\begin{tcolorbox}[promptbox,title={Iterative Refinement Prompt (Augmented Generation)}]
You are given the annotation for a video.

\textbf{Taxonomy JSON:}

\texttt{\textless taxonomy json\textgreater}

\textbf{Task:}
Create exactly one difficult, hint-free question for evaluating a video understanding model.

\textbf{Hard constraints:}

- The question must NOT contain words that leak the target semantic field, such as:
  \texttt{hate}, \texttt{bias}, \texttt{stereotype}, \texttt{discrimination}, \texttt{offensive}, \texttt{toxic}, \texttt{negative}, \texttt{sarcasm}, \texttt{irony}, \texttt{prejudice}, \texttt{controversial}

- The question should not presuppose that the video has hidden meaning.

- The question should require understanding the intended meaning, not just literal description.

- The gold answer should be concise but semantically complete.

- The rubric should make later LLM judging reliable.

The previous QA attempt was judged as:

\texttt{\textless validation json\textgreater}

You must improve the QA package based on this feedback.

\textbf{Important improvement instructions:}

- Fix the issues listed in \texttt{"issues"} and \texttt{"fix\_suggestions"}.

- If \texttt{"surface\_answerable"} was true, make the new question require a more specific implicit inference.

- If \texttt{"requires\_implicit\_understanding"} was false, make the question force recovery of the intended meaning.

- If \texttt{"difficulty\_fit"} was \texttt{"too\_easy"} or \texttt{"too\_hard"}, recalibrate the question difficulty.

- If a \texttt{"revised\_question"} is provided by the validator, you may adopt or improve it.

- Keep the question neutral and non-leading.
\end{tcolorbox}

This prompt is used during iterative refinement rounds. 

\begin{tcolorbox}[promptbox,title={Multi-Choice Question Prompt}]
You are answering a multi-choice question about a video.

\textbf{Question:}

\texttt{\textless question\textgreater}

\textbf{Transcript from ASR (may be noisy, partial, or empty):}

\texttt{\textless transcript\textgreater}

\texttt{\textless extra\_context\textgreater}

\textbf{Instruction:}

\texttt{\textless instruction\textgreater}

\textbf{Available options:}

\texttt{\textless option\_text\textgreater}

Return only valid JSON with one field:

\texttt{\{\{}\\
\texttt{\ \ "selected\_options": ["A", "B"]}\\
\texttt{\}\}}

\textbf{Rules:}

- Select only options that are clearly supported by the video.

- Return only option letters, not option texts.

- Do not include any option not in the provided list.

- Be conservative: do not select an option unless it is clearly justified by the video.

- If no option is clearly supported, return an empty list.
\end{tcolorbox}

\begin{tcolorbox}[promptbox,title={Open-Ended Question Prompt}]
You are answering a question about a video.

\textbf{Question:}

\texttt{\{question\}}

\textbf{Transcript from ASR (may be noisy, partial, or empty):}

\texttt{\{transcript\}}

\textbf{Instructions:}

- Answer the question directly.

- Use the video frames as the primary source of truth.

- Infer visible on-screen text from the frames when relevant.

- Keep the answer concise but semantically complete.

- Do not add safety disclaimers unless absolutely necessary.

\end{tcolorbox}

\section{LLM-as-a-Judge for Open-Ended Questions}
\label{app_laaj}

To evaluate model performance on open-ended, hint-free questions, we adopt an LLM-as-a-judge framework that assesses semantic understanding rather than surface-level similarity. Instead of relying on exact match or n-gram overlap, the judge model evaluates whether a prediction captures the intended meaning of the video.

\paragraph{Judging Framework.}
Given a question $q$, a gold answer $a^*$, and a model prediction $\hat{a}$, the judge receives the following structured inputs:
(i) the question,
(ii) the gold answer,
(iii) a set of reference points summarizing key aspects of the intended meaning, and
(iv) a grading rubric specifying evaluation criteria.
The judge then produces a structured judgment consisting of a scalar score, a detailed breakdown, and a qualitative verdict.

\paragraph{Scoring Dimensions.}
The evaluation decomposes semantic understanding into five components:

\begin{itemize}
    \item \textbf{Core Intent} ($0$--$5$): whether the prediction captures the video's primary intended meaning.
    \item \textbf{Implicit Signal} ($0$--$3$): whether it correctly identifies the key rhetorical or social signal.
    \item \textbf{Target or Social Meaning} ($0$--$1$): whether it recognizes relevant targets, groups, or social implications when applicable.
    \item \textbf{Hallucination Penalty} ($0$--$3$): penalizes unsupported or fabricated claims.
    \item \textbf{Literal-Only Penalty} ($0$--$3$): penalizes answers that remain at surface-level description without capturing subtext.
\end{itemize}

The final score is computed as:
\begin{equation}
\begin{aligned}
\text{score}_{\text{total}} =
&\ \text{core\_intent}
+ \text{implicit\_signal}
+ \text{target\_or\_social\_meaning} \\
&- \text{hallucination\_penalty}
- \text{literal\_only\_penalty}.
\end{aligned}
\end{equation}

This formulation explicitly rewards semantic understanding while penalizing both hallucination and shallow interpretation.

\paragraph{Judgment Output.}
In addition to the scalar score, the judge produces:
(i) a structured score breakdown,
(ii) a categorical verdict from \{excellent, good, partial, poor, wrong\}, and
(iii) a short natural language justification.

This structured output enables both quantitative evaluation and qualitative analysis of model behavior.

\paragraph{Design Principles.}
The LLM judge is designed to follow three principles:
\begin{itemize}
    \item \textbf{Semantic over lexical matching:} evaluations are based on meaning rather than wording.
    \item \textbf{Strict hallucination control:} unsupported claims are explicitly penalized.
    \item \textbf{Subtext sensitivity:} answers that fail to capture implicit meaning receive lower scores even if they are factually correct at the surface level.
\end{itemize}

\begin{tcolorbox}[promptbox,title={Judge Prompt}]
You are grading answers in a benchmark for hint-free implicit video understanding.

Your job is to judge \textbf{semantic understanding}, not style.

\textbf{Scoring dimensions:}

- \texttt{core\_intent}: Did the model capture the main intended meaning?

- \texttt{implicit\_signal}: Did it recognize the crucial hidden rhetorical or social signal?

- \texttt{target\_or\_social\_meaning}: Did it identify a relevant target, group, institution, or social implication when supported?

- \texttt{hallucination\_penalty}: Penalize invented claims not grounded by the gold answer, evidence, or rubric.

- \texttt{literal\_only\_penalty}: Penalize answers that remain at surface description and miss the point.

\textbf{Score bounds:}

- \texttt{core\_intent} must be one of {0, 1, 2, 3, 4, 5}

- \texttt{implicit\_signal} must be one of {0, 1, 2, 3}

- \texttt{target\_or\_social\_meaning} must be one of {0, 1}

- \texttt{hallucination\_penalty} must be one of {0, 1, 2, 3}

- \texttt{literal\_only\_penalty} must be one of {0, 1, 2, 3}

\textbf{Scoring rule:}

\texttt{score\_total} =

\texttt{core\_intent}
+ \texttt{implicit\_signal}
+ \texttt{target\_or\_social\_meaning}

- \texttt{hallucination\_penalty}
- \texttt{literal\_only\_penalty}

The maximum possible score is 9.

\textbf{Interpretation guide:}

- A partially correct answer that captures the meme's point should score much higher than a polished but purely literal answer.

- Do not require exact wording match.

- Be strict with hallucinations.

- Only assign \texttt{target\_or\_social\_meaning = 1} when that dimension is genuinely relevant and correctly captured.

- Use the evidence as grounding support, not as extra hidden labels to overfit.

- Keep \texttt{reasoning\_short} concise.

Return only \textbf{valid JSON}.
\end{tcolorbox}

\section{Rhetoric Mechanism Grouping}
\label{app_rmgroup}

Table~\ref{tab:rhetoric_macro_mapping} defines the mapping from fine-grained rhetoric mechanism labels to five macro categories, which serve as the basis for structured evaluation and analysis.

\section{Social Value Signals Grouping}
\label{app_svgroup}

Table~\ref{tab:social_macro_mapping} defines the mapping from fine-grained social value signals to macro-level categories, enabling consistent evaluation of subtext across models.

\section{Details of the Taxonomy of Rhetoric Mechanisms}
\label{tax_rm}

\paragraph{Literal Only.} The video’s meaning is largely exhausted by its surface content, with little or no reliance on non-literal interpretation, rhetorical reframing, or implicit contrast.

\begin{wraptable}{r}{0.55\textwidth}
\centering
\vspace{-1em}
\caption{Mapping for rhetoric mechanisms.}
\label{tab:rhetoric_macro_mapping}
\small
\begin{tabular}{ll}
\toprule
\textbf{Macro Category} & \textbf{Subcategory} \\
\midrule
Literal / Direct & Literal Only \\

Opposition / Incongruity & Contrast \\
 & Bait and Switch \\
 & Role Reversal \\
 & Absurdism \\

Attitude / Tone-based Rhetoric & Sarcasm \\
 & Irony \\
 & Deadpan \\
 & Mockery \\

Amplification / Stylization & Exaggeration \\
 & Parody \\

Implicit / Coded Social Framing & Innuendo \\
 & Stereotype Invocation \\
 & Dog Whistle or Code \\
 & Other \\
\bottomrule
\end{tabular}
\vspace{-5em}
\end{wraptable}
\paragraph{Sarcasm.} The video conveys meaning by expressing a surface attitude that is intentionally opposite to the speaker’s or creator’s actual attitude, typically to signal ridicule, dismissal, or criticism.

\paragraph{Irony.} The video derives meaning from a discrepancy between appearance and reality, expectation and outcome, or explicit expression and underlying implication, without necessarily requiring direct mocking intent.

\paragraph{Mockery.} The video is structured to ridicule, belittle, or make fun of a target, often by highlighting flaws, incompetence, absurdity, or hypocrisy.

\paragraph{Stereotype Invocation.} The video relies on a recognizable stereotype, trope, or socially shared caricature to construct its meaning, whether for humor, critique, reinforcement, or inversion.

\paragraph{Exaggeration.} The intended meaning is amplified through deliberate overstatement, extreme depiction, or disproportionate framing beyond what would be literally plausible.

\paragraph{Contrast.} The meaning is produced through juxtaposition between two incompatible or sharply different elements, such as tone, image, text, expectation, or social role.

\paragraph{Innuendo.} The video implies a sensitive, suggestive, or socially loaded meaning indirectly, without stating it explicitly, often relying on implication rather than overt expression.

\paragraph{Absurdism.} The video constructs meaning through deliberate irrationality, impossibility, or surreal mismatch, where the humor or point depends on embracing the nonsensical.

\paragraph{Role Reversal.} The video derives meaning by inverting expected roles, positions, hierarchies, or behavioral norms, so that one party acts in a way conventionally associated with another.

\paragraph{Dog Whistle or Code.} The video contains coded references, euphemisms, or indirect signals that are intended to be legible primarily to audiences with relevant cultural, political, or subcultural knowledge.

\begin{wraptable}{r}{0.5\textwidth}
\centering
\caption{Mapping for social value signals.}
\label{tab:social_macro_mapping}
\small
\resizebox{\linewidth}{!}{%
\begin{tabular}{ll}
\toprule
\textbf{Macro Category} & \textbf{Subcategory} \\
\midrule
Neutral / No Social Signal & None \\

Emotional Attitude & Negative Affect \\
 & Fatalism or Cynicism \\

Social Evaluation / Devaluation & Contempt \\
& Humiliation \\
& Aggression or Hostility \\
& Exclusion \\
& Discrimination or Prejudice \\

Norm and Value Framing & Norm Violation \\
 & Anti-mainstream Value \\

Identity / Ideological Signaling & Political or Identity Signal \\
 & Sexual Implication \\
 & Other \\
\bottomrule
\end{tabular}%
}
\vspace{-5em}
\end{wraptable}
\paragraph{Bait and Switch.} The video sets up one expectation and then abruptly replaces it with a different, often incompatible, payoff, producing humor or commentary through misdirection.

\paragraph{Deadpan.} The video presents absurd, ironic, or exaggerated content in a deliberately flat, matter-of-fact, or emotionally neutral manner, making the restrained delivery central to the effect.

\paragraph{Parody.} The video imitates the style, structure, or conventions of another genre, person, discourse, or media form in order to create humor, critique, or commentary.

\paragraph{Other.} The video relies on a rhetorical mechanism not adequately captured by the categories above, or on a hybrid mechanism that cannot be cleanly reduced to a single listed type.

\section{Details of the Taxonomy of Social Value Signals}
\label{tax_sv}

\paragraph{None.} The video does not clearly communicate a salient social attitude, value judgment, or normative stance beyond its immediate surface content.

\paragraph{Negative Affect.} The video conveys or evokes a broadly negative emotional tone, such as frustration, displeasure, discomfort, annoyance, or aversion, without necessarily specifying a stronger social stance.

\paragraph{Contempt.} The video signals scorn, disdain, or a sense of superiority toward a target, often implying that the target is foolish, inferior, pathetic, or unworthy of respect.

\paragraph{Exclusion.} The video communicates boundary-making, rejection, or denial of belonging, whether socially, culturally, morally, or group-wise.

\paragraph{Discrimination or Prejudice.} The video conveys bias, derogation, or unequal judgment toward a group or identity category, whether explicitly or through implication, stereotype, or coded framing.

\paragraph{Norm Violation.} The video foregrounds behavior, values, or situations as improper, transgressive, taboo, or outside expected social rules or conventions.

\paragraph{Anti-Mainstream Value.} The video endorses, celebrates, or signals attitudes positioned against widely accepted norms, tastes, or mainstream moral or social expectations.

\paragraph{Fatalism or Cynicism.} The video expresses resignation, hopelessness, distrust, or a dismissive belief that outcomes, people, or institutions are fundamentally flawed or unchangeable.

\paragraph{Sexual Implication.} The video conveys sexualized meaning, innuendo, erotic framing, or sexually suggestive interpretation, whether humorous, implicit, or socially coded.

\paragraph{Political or Identity Signal.} The video communicates a political stance, ideological alignment, or identity-linked signal, including cues tied to collective affiliation, social positioning, or worldview.

\paragraph{Aggression or Hostility.} The video conveys antagonism, threat, attack, intimidation, or overtly adversarial attitude toward a target.

\paragraph{Humiliation.} The video frames a person or target as embarrassed, degraded, exposed, or socially diminished, often making loss of dignity central to the effect.

\paragraph{Other.} The video communicates a social attitude or value signal not adequately captured by the categories above, or one that combines multiple signals without a clear dominant type.

\section{Guided and Unguided Prompts for Structured Subtext Understanding Tasks}
\label{app_guided_prompts}

For the two structured subtext understanding tasks, namely rhetoric mechanism identification and social value signal identification, we evaluate models under two prompt settings: \textit{without guidance} and \textit{with guidance}. The unguided setting provides only the task question, transcript, options, and output rules, while the guided setting additionally provides taxonomy definitions for the five macro categories.

\begin{tcolorbox}[promptbox,title={Without-Guidance Prompt}]
You are answering a multi-choice question about a video.

\textbf{Task name:}

\texttt{\textless task\_name\textgreater}

\textbf{Question:}

\texttt{\textless question\textgreater}

\textbf{Transcript from ASR} (may be noisy, partial, or empty):

\texttt{\textless transcript\textgreater}

\textbf{Instruction:}

\texttt{\textless instruction\textgreater}

\textbf{Available options:}

\texttt{\textless option\_text\textgreater}

Return only valid JSON with one field:

\{
  "\texttt{selected\_options}": ["A", "B"]
\}

\textbf{Rules:}

- Select only options that are clearly supported by the video.

- Return only option letters, not option texts.

- Do not include any option not in the provided list.

- Be conservative: do not select an option unless it is clearly justified by the video.

- If no option is clearly supported, return an empty list.
\end{tcolorbox}

\begin{tcolorbox}[promptbox,title={With-Guidance Prompt}]
You are analyzing a video and answering a multi-choice question.

Your goal is to identify the most appropriate categories based on the video's meaning.

\textbf{Task name:}

\texttt{\textless task\_name\textgreater}

\textbf{Question:}

\texttt{\textless question\textgreater}

\textbf{Transcript from ASR} (may be noisy, partial, or empty):

\texttt{\textless transcript\textgreater}

\texttt{\textless taxonomy\_guidance\textgreater}

\textbf{Instruction:}

\texttt{\textless instruction\textgreater}

\textbf{Available options:}

\texttt{\textless option\_text\textgreater}

\textbf{Output format:}

Return ONLY valid JSON with the following structure:

\{
  "\texttt{selected\_options}": ["A", "B"]
\}

\textbf{Rules:}

- Only return option letters (A--E).

- Do NOT return option texts.

- Select all options that are clearly supported by the video.

- Do NOT guess if evidence is weak.

- If none apply, return an empty list.
\end{tcolorbox}

\begin{tcolorbox}[promptbox,title={Rhetoric Guidance}]
Additional guidance on taxonomy categories:

The options correspond to \textbf{high-level rhetorical categories}.
Each category summarizes several finer-grained rhetorical patterns commonly observed in video memes.

\textbf{Rhetoric Macro Categories:}

\textbf{A. Literal / Direct}

Meaning is conveyed directly without rhetorical transformation.

Typical patterns include:

- \texttt{literal\_only}: the video communicates its message directly without irony, exaggeration, or figurative framing.

\textbf{B. Opposition / Incongruity}

Meaning arises from contradiction, reversal, or unexpected juxtaposition.

Typical patterns include:

- \texttt{contrast}: juxtaposing two opposing situations, ideas, or outcomes.

- \texttt{bait\_and\_switch}: setting up one expectation and then suddenly replacing it with a different or contradictory outcome.

- \texttt{role\_reversal}: reversing expected roles, identities, or positions to produce humor or commentary.

- \texttt{absurdism}: presenting illogical or exaggerated situations that highlight incongruity.

\textbf{C. Attitude / Tone-based Rhetoric}

Meaning is conveyed primarily through tone or speaker attitude.

Typical patterns include:

- \texttt{sarcasm}: expressing a meaning by stating the opposite of what is intended.

- \texttt{irony}: the intended meaning contrasts with the literal situation or appearance.

- \texttt{deadpan}: presenting absurd or humorous content in a serious, emotionless manner.

- \texttt{mockery}: ridiculing or making fun of a person, behavior, or situation.

\textbf{D. Amplification / Stylization}

Meaning is emphasized through exaggeration or stylized imitation.

Typical patterns include:

- \texttt{exaggeration}: overstating a situation or characteristic to emphasize its significance.

- \texttt{parody}: imitating the style or conventions of a person, genre, or cultural artifact for humorous or critical effect.

\textbf{E. Implicit / Coded Social Framing}

Meaning is conveyed through indirect or socially coded signals.

Typical patterns include:

- \texttt{innuendo}: suggesting a meaning indirectly rather than stating it explicitly.

- \texttt{stereotype\_invocation}: referencing widely known stereotypes to imply a social meaning.

- \texttt{dog\_whistle\_or\_code}: using coded expressions that convey specific meanings to certain audiences while remaining subtle to others.

\textbf{When answering:}

- Focus on how the video constructs its underlying meaning.

- Identify the rhetorical strategies used to convey humor, critique, or commentary.

- A video may contain multiple rhetorical mechanisms.
\end{tcolorbox}

\begin{tcolorbox}[promptbox,title={Social Value Guidance}]
Additional guidance on taxonomy categories:

The options correspond to \textbf{high-level social value signals}.
Each category summarizes several finer-grained social attitudes or stances that may be expressed in video memes.

\textbf{Social Value Signal Categories:}

\textbf{A. Neutral / No Social Signal}

The video conveys humor or content without expressing a clear social stance.

Typical patterns include:

- \texttt{none}: the video does not convey a noticeable social judgment, stance, or value signal.

\textbf{B. Emotional Attitude}

The video expresses a general emotional tone or affect toward a situation.

Typical patterns include:

- \texttt{negative\_affect}: expressing frustration, disappointment, annoyance, or dissatisfaction.

- \texttt{fatalism\_or\_cynicism}: expressing pessimism, resignation, or cynical attitudes about situations or outcomes.

\textbf{C. Social Evaluation / Devaluation}

The video evaluates, criticizes, or demeans people or groups.

Typical patterns include:

- \texttt{contempt}: expressing disdain or disrespect toward someone or something.

- \texttt{humiliation}: portraying someone as foolish, incompetent, or inferior.

- \texttt{aggression\_or\_hostility}: showing hostility, threats, or aggressive attitudes.

- \texttt{exclusion}: implying that certain people or groups should be excluded or marginalized.

- \texttt{discrimination\_or\_prejudice}: expressing biased or discriminatory attitudes toward social groups.

\textbf{D. Norm and Value Framing}

The video comments on social rules, expectations, or cultural norms.

Typical patterns include:

- \texttt{norm\_violation}: highlighting or mocking behavior that breaks accepted social rules.

- \texttt{anti\_mainstream\_value}: expressing opposition to widely accepted social norms or values.

\textbf{E. Identity / Ideological Signaling}

The video references identity, ideology, or social group affiliation.

Typical patterns include:

- \texttt{political\_or\_identity\_signal}: expressing political stances or identity-based perspectives.

- \texttt{sexual\_implication}: implying sexual themes or identity-related meanings.

- \texttt{other}: conveying social signals that do not clearly fall into the above categories.

\textbf{When answering:}

- Focus on what social stance, value judgment, or attitude the video conveys.

- Identify whether the video expresses opinions about people, groups, norms, or identities.

- A video may express multiple social value signals.
\end{tcolorbox}

\section{The With-Guidance Counterpart of Affinity Bias}
\label{app_wguidance_fig9}

\begin{figure}[t]
    \centering
    \begin{subfigure}[t]{0.48\textwidth}
        \centering
        \includegraphics[width=\linewidth]{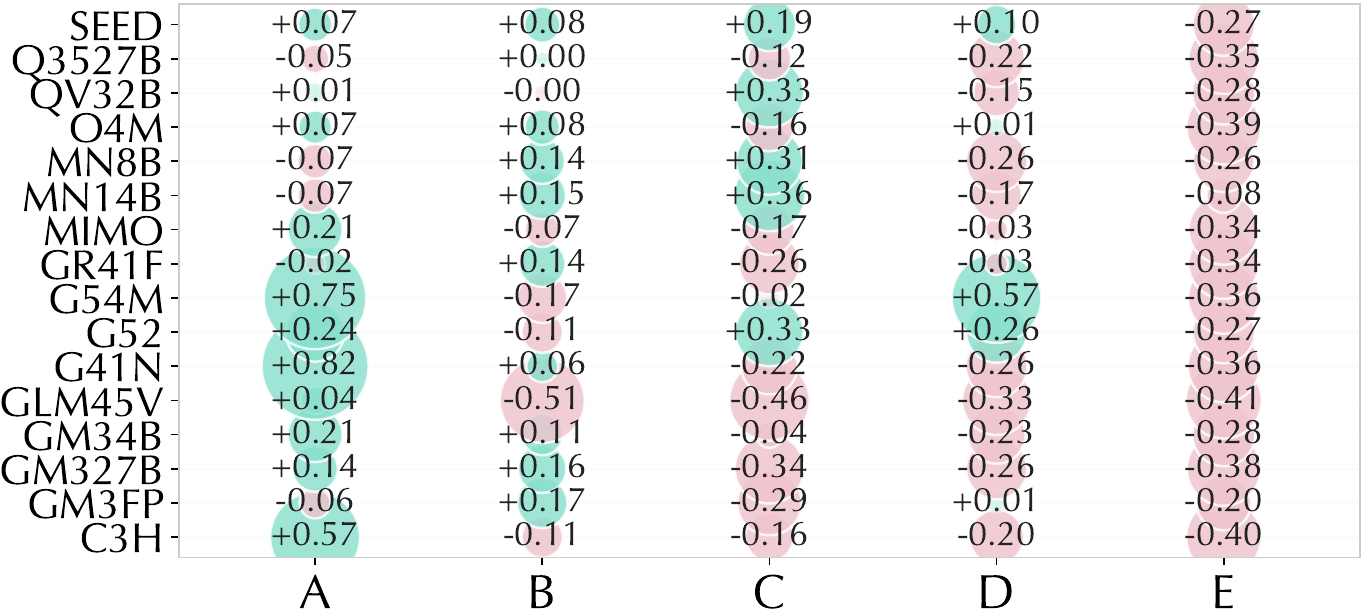}
        \caption{Rhetoric Mechanisms Identification.}
        \label{fig:affinity_rhetoric_app}
    \end{subfigure}
    \hfill
    \begin{subfigure}[t]{0.48\textwidth}
        \centering
        \includegraphics[width=\linewidth]{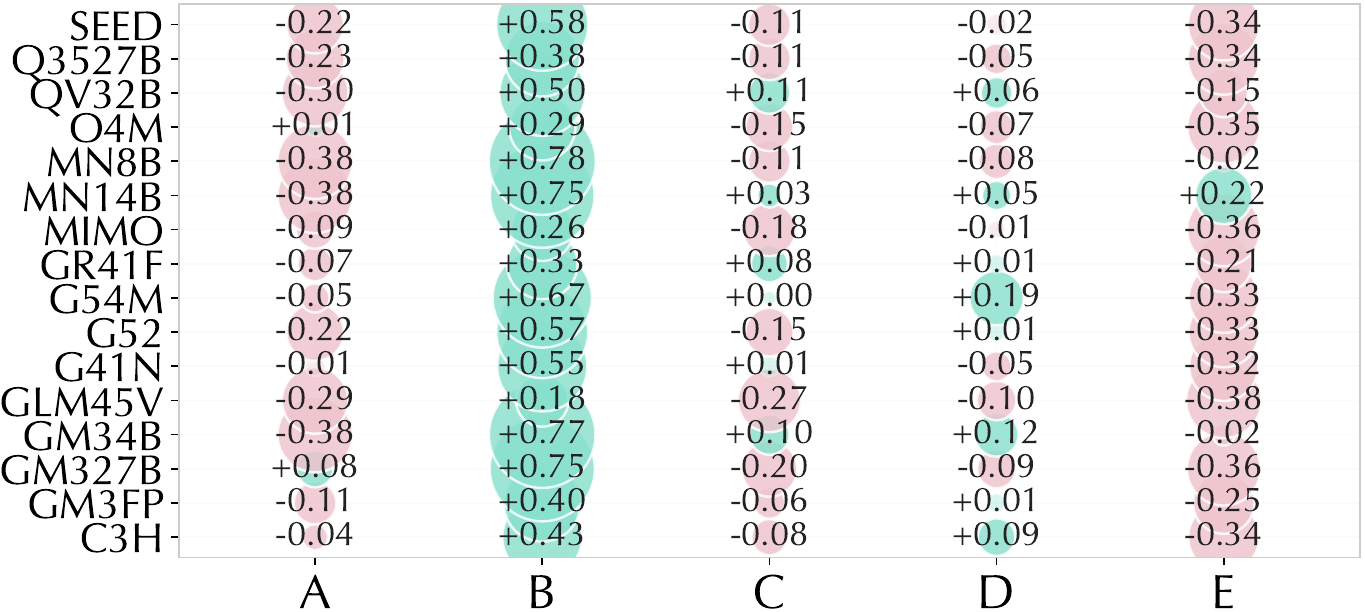}
        \caption{Social Value Signal Identification.}
        \label{fig:affinity_social_app}
    \end{subfigure}
    \vspace{-2mm}
    \caption{Model--option affinity bias with guidance. Positive values indicate over-prediction relative to ground-truth prevalence, while negative values indicate under-prediction.}
    \label{fig:affinity_maps_app}
    \vspace{-4mm}
\end{figure}

The with-guidance results related to Figure~\ref{fig:affinity_maps_main} is given in Figure~\ref{fig:affinity_maps_app}. As can be observed, the results of these two figures show simiular pattern.

\section{Details of Baselines and the Evaluation Process}
\label{app_eval_detial}

For all tasks, model outputs are normalized into percentage scores for consistent comparison across different evaluation settings.

For the open-ended (OE) task, we adopt an LLM-as-a-judge evaluation protocol (as described in Appendix~\ref{app_laaj}). Each prediction is compared against the reference answer using a structured rubric that evaluates whether the model captures the core intended meaning, key implicit signals, and relevant social interpretation, while penalizing hallucination and purely literal responses. The final score is aggregated into a continuous value in $[0,1]$ and then scaled to a percentage.

For the multiple-choice tasks (including evidence grounding, rhetoric mechanism identification, and social value signal identification), we adopt a strict yet interpretable set-based scoring rule. Let $\mathcal{P}$ denote the set of predicted options and $\mathcal{G}$ denote the ground-truth set. If the prediction contains any incorrect option (i.e., $\mathcal{P} \setminus \mathcal{G} \neq \emptyset$), the score is assigned as $0$. Otherwise, the score is computed as the proportion of correctly selected options:
\[
\text{score} = \frac{|\mathcal{P} \cap \mathcal{G}|}{|\mathcal{G}|}.
\]
This design ensures that models are penalized for hallucinated selections, while still receiving partial credit when they correctly identify a subset of the required evidence or categories. A full score is only obtained when all and only the correct options are selected.

Finally, task-level scores are averaged across all samples, and overall performance is reported as the mean across tasks.

\section{Limitations}

Despite its broad coverage of subtext understanding, ViMU has several limitations. First, the interpretation of metaphorical and socially grounded meaning is inherently subjective, and although we employ structured annotation and validation procedures, residual ambiguity and annotator bias may remain. While ViMU is designed for evaluation rather than training, models may still exploit superficial patterns or dataset-specific regularities, and strong performance on this benchmark does not necessarily imply robust real-world understanding of nuanced social or cultural meaning. Overall, these limitations reflect broader challenges in constructing benchmarks for subjective and socially situated understanding, rather than weaknesses unique to ViMU.

\section{Societal Impact}

ViMU aims to advance the evaluation of multimodal models by focusing on their ability to interpret implicit, socially grounded meanings in videos. A positive impact of this work is that it helps expose systematic limitations of current models in understanding rhetoric, social signals, and culturally situated subtext, which are critical for safe and reliable deployment in real-world applications such as content moderation, assistive technologies, and human--AI interaction. 

However, the dataset also involves potential risks. Because it includes socially sensitive and potentially offensive content, there is a possibility that models evaluated on ViMU may reproduce or amplify harmful stereotypes, biases, or misinterpretations. In addition, improved capability in interpreting implicit meaning could be misused for profiling, surveillance, or manipulation of user intent, especially in contexts involving political or identity-related signals. There is also a risk that benchmark performance may be overinterpreted as a proxy for real-world social understanding, despite the inherent subjectivity and cultural dependency of such tasks.

We want to emphasize that ViMU is intended \textbf{solely as an evaluation benchmark rather than a training resource}, and we encourage users to carefully consider the dataset's limitations, report model behaviors transparently, and avoid deploying systems based solely on benchmark performance. Future work should further investigate fairness, cultural coverage, and robustness to ensure that advances in subtext understanding benefit diverse user groups without reinforcing existing harms.

\end{document}